\documentclass[11pt]{article}
\usepackage[margin=1in]{geometry}
\usepackage{setspace}
\usepackage{amsmath, amssymb, amsthm}
\usepackage{mathtools}
\usepackage{graphicx}
\usepackage[font=small,labelfont=bf]{caption}
\usepackage[authoryear,round]{natbib}  
\usepackage{hyperref}
\hypersetup{
    colorlinks=true,
    linkcolor=black,
    citecolor=black,
    urlcolor=black
}
\usepackage{algorithm}
\usepackage{algorithmic}
\usepackage{booktabs}
\usepackage{multirow}
\usepackage{longtable}

\theoremstyle{definition}

\theoremstyle{remark}

\usepackage{enumerate}
\usepackage{url}

\title{How many stations are sufficient? \\Exploring the effect of urban weather station density reduction on imputation accuracy of air temperature and humidity.}

\author{
    Marvin Plein$^{1}$, Carsten F. Dormann$^{1}$, Andreas Christen$^{2}$ \\[0.3em]
    $^{1}$Biometry and Environmental System Analysis, University of Freiburg, Germany \\
    $^{2}$Chair of Environmental Meteorology, University of Freiburg, Germany \\[0.3em]
    Email: \texttt{marvin.plein@biom.uni-freiburg.de}
}

\date{\today}
\begin{document}
\maketitle

\begin{abstract}
Urban weather station networks (WSNs) are widely used to monitor urban weather and climate patterns and aid urban planning. However, maintaining WSNs is expensive and labor-intensive. Here, we present a step-wise station removal procedure to thin an existing WSN in Freiburg, Germany, and analyze the ability of WSN subsets to reproduce air temperature and humidity patterns of the entire original WSN for a year following a simulated reduction of WSN density.
We found that substantial reductions in station numbers after one year of full deployment are possible while retaining high predictive accuracy. A reduction from 42 to 4 stations, for instance, increased mean prediction RMSEs from 0.69 K to 0.83 K for air temperature and from 3.8\% to 4.4\% for relative humidity, corresponding to RMSE increases of only 20\% and 16\%, respectively. Predictive accuracy is worse for remote stations in forests than for stations in built-up or open settings, but consistently better than a state-of-the-art numerical urban land-surface model (Surface Urban Energy and Water Balance Scheme). Stations located at the edges between built-up and rural areas are most valuable when reconstructing city-wide climate characteristics. 
Our study demonstrates the potential of thinning WSNs to maximize the efficient allocation of financial and personnel-related resources in urban climate research.
\end{abstract}

\maketitle

\section{Introduction}
In recent years, several cities established high-resolution weather station networks (WSNs). Such networks usually measure urban-rural and intra-urban differences in atmospheric variables during both typical and extreme weather conditions. For example, air temperature ($T_a$), humidity, sometimes also wind and mean radiant temperatures are used in a wide range of planning and management decisions in urban areas, including the development of heat stress mitigation strategies \citep{oke2017urban, ward2016heat, gubler2021evaluation} and the assessment of building energy demand \citep{abunnasr2022sebu, mirzaei2010approaches}. WSNs are also helpful in providing near-time information on extreme events, such as heat, heavy precipitation associated with flooding, or windstorms. Due to technological advances \citep{perera2014sensing} and the increased availability of reliable and networked mid-cost sensors \citep{honjoNetworkOptimizationEnhanced2015}, urban WSNs have become more available for capturing, in particular, intra-urban $T_a$ and humidity patterns \citep[e.g.][]{chang2010development, basara2011oklahoma, chapman2015birmingham, cecilia2023measuring, plein2025gapfilling, feigel2025high}.

Given the initial installation and continuing operating costs and the administrative efforts of planning and implementing a WSN \citep{muller2013sensors}, it is key to identify the optimal spatial density and distribution of stations for given available resources, the specific research purposes (e.g., analysis of single heat wave or long-term urban climate patterns) and external constraints (availability and distribution of accessible and authorized measurement locations, spatial patterns of urban structure, administrative responsibilities, etc.) \citep{schlunzen2023guidance}. 
However, also during the deployment of a WSN, there are important issues to be addressed. Over time, the frequency of both sensor malfunctions and vandalism or theft increases, necessitating decisions on whether to replace the corresponding stations or relinquish a measurement location, as future information gain is limited. Apart from such forced considerations, research by \cite{plein2025gapfilling} shows that a deployment time of one year is sufficient to use data from proximate stations to obtain accurate estimates of $T_a$ and relative humidity ($RH$) at specific stations using learned patterns of similarity between measurement locations. This suggests the option of reducing or rearranging a WSN after an initial period of full deployment, while retaining the detailed spatiotemporal information that a dense network provides. 

Despite the growing number of WSNs deployed in cities globally, there is little research on improving WSN efficiency by removing redundant stations. \citet{honjoNetworkOptimizationEnhanced2015} explore the effects of WSN density reduction based on random station sampling and hierarchical clustering (using geographic coordinates) on generating interpolated maps of $T_a$ in a WSN of 200 stations in the Tokyo metropolitan area \citep{yamato2009new}. \citet{dingMachineLearningassistedMapping2023} expand this research by additionally including land cover for clustering and comparing the interpolation accuracy of eight types of interpolation models, including classical spatial interpolation algorithms, regression models, and combinations of the two modeling approaches in a WSN encompassing 321 stations in Guangzhou, China. 

While both studies found that substantial reductions in WSN density could be achieved with minimal deterioration in interpolation accuracy, they leave room for further research in several key aspects. First, both studies are limited to $T_a$. Given that most current WSNs at least also record humidity, WSN optimization should ideally be performed with consideration of both variables, as demonstrated by \citet{chenUrbanClimateMonitoring2022}. Second, the periods used to evaluate the WSN reduction effects fell within the deployment time of the full network. Since a real-world application of WSN reduction would involve operating a thinned network in the future, we argue that the effects of WSN reduction should be evaluated over a period following the (hypothetical) removal of stations, corresponding to temporal extrapolation. Lastly, both studies evaluate the performance of reduced WSNs by comparing spatial grids of interpolated data, namely those generated by the full network, which serve as ground truth, and those produced by thinned WSNs. In our view, this is problematic as the full network interpolations already contain interpolation errors, which makes it difficult to quantify the prediction errors resulting from the interpolation on the one hand and the WSN density reduction on the other hand.

This study aims to address these issues by simulating a density reduction of a currently operative WSN consisting of 42 stations located in the city and vicinity of Freiburg, Germany. Using data from the first year of continuous operation, we train machine-learning models (Extreme Gradient Boosting, EGB   \citep{xgboost}) to capture the spatio-temporal relationships between $T_a$ and humidity, expressed as vapor pressure ($e$), between all stations. Then, we apply a data-driven step-wise station removal procedure that iteratively selects stations to remove from the network, optimized to minimize imputation errors across all stations and variables simultaneously. This results in subsets of retained stations, which are then used to train models that reproduce the data of all stations from the second operational year. The performance of models trained on identified station subsets of varying size is evaluated against several reference models, namely predictions generated by a simple Generalized Linear Model (GLM) trained on data from two official weather stations situated in the study area, results of a numerical urban land surface model  \citep[Surface Urban Energy and Water Balance Scheme (SUEWS):][]{jarvi2011surface, ward2016surface, sun2019python}, and predictions of EGB  models trained on randomly selected station subsets, and on both years of data.

\section{Methods}
\label{Methods}

\subsection{Study area and period}

\begin{figure}[ht!]
 \centerline{\includegraphics[width=39pc]{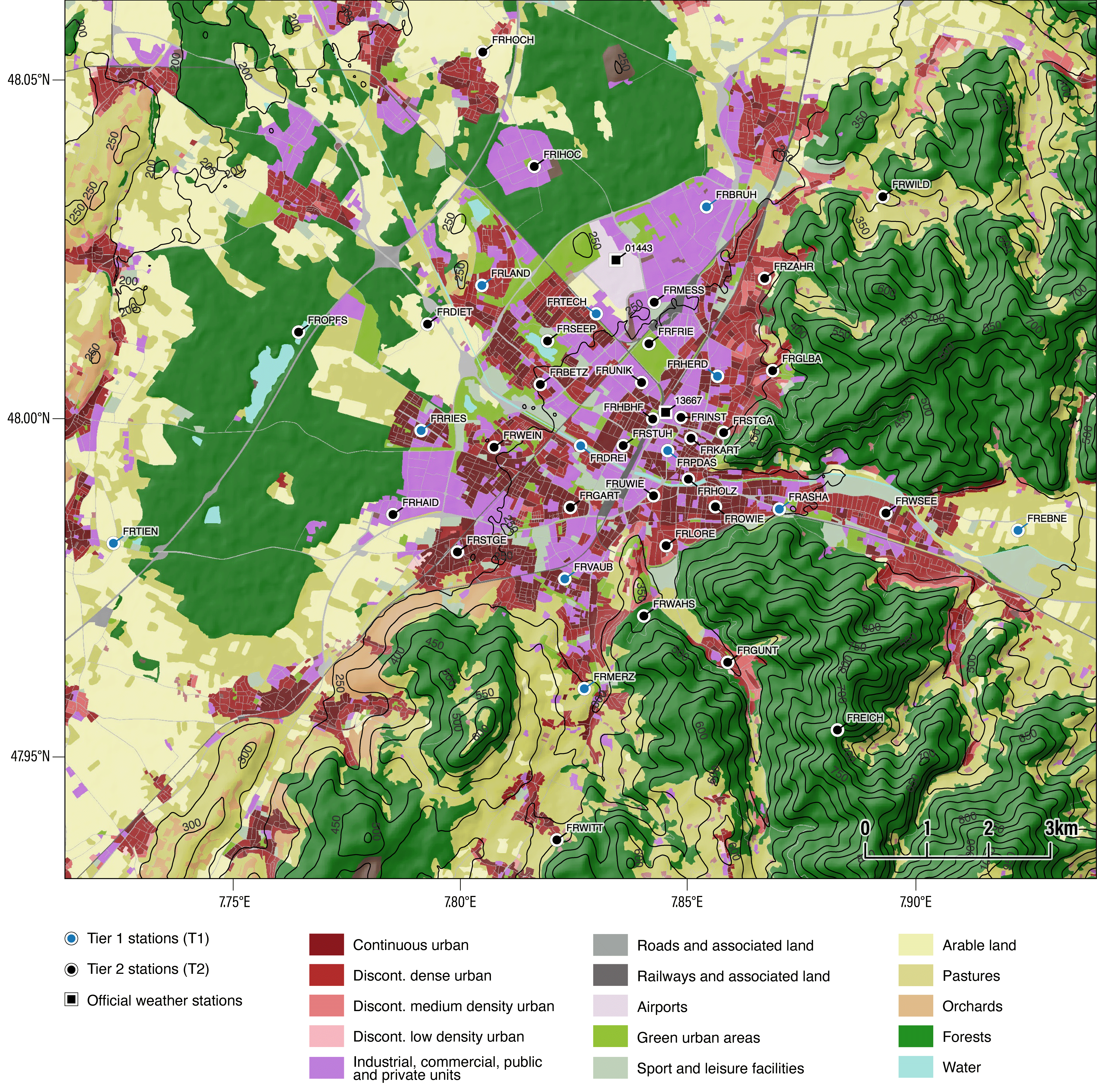}}
  \caption{Overview of the study area and the weather station locations. Figure taken from \cite{plein2025gapfilling}.}
  \label{study_area}
  \label{fig:study_area}
\end{figure}

Freiburg im Breisgau is a mid-size city (236,000 inhabitants) situated in southwestern Germany (48°00'~N, 07°51'~E) (Statistisches Landesamt, 2022). The administrative area of the city covers approximately 150~km$^2$ in the southeastern Upper Rhine Valley with the Black Forest mountains located immediately east of the city (Fig.~\ref{fig:study_area}). While exhibiting a temperate oceanic climate \citep[Cfb, according to the Köppen-Geiger climate classification:][]{koppen1931grundriss}, its complex orography results in thermally driven mountain-valley wind patterns and complex local-scale weather patterns, particularly in the southern and eastern parts of the city \citep{ernst1995tagesperiodische, gross1989numerical, rockle2003klimaanalyse, nubler1979konfiguration}. 

The study period spanned from September 1, 2022, to August 31, 2024, with the first year allocated for training and the second year for evaluation purposes. The second year was both warmer and wetter than the first year, exhibiting a mean $T_a$ of 13.2~°C and a precipitation sum of 1013 mm compared to 12.6~°C and 780 mm (Fig. \ref{fig:combined}), calculated using data from the official weather station 01443 operated by the German Meteorological Service (Deutscher Wetterdienst, DWD) (see Fig.~\ref{fig:study_area}).

\subsection{Data}

\subsubsection*{WSN data and pre-processing}

The WSN used in this study was installed between March and November 2022 and consists of 42 stations located in both built-up and non-built-up areas of the city (Fig \ref{fig:study_area}). 13 Tier 1 (T1) stations feature a ClimaVue50 sensor manufactured by Campbell Scientific, while at the remaining 29 locations a nMetos180 weather station by Pessl Instruments was installed. A detailed description of the WSN characteristics is provided by \cite{plein2025gapfilling}, while \cite{feigel2025high} focus on the more sophisticated T1 stations. Stations are labeled using a six-letter code, in which the first two letters (``FR'') indicate the city (Freiburg) and the remaining four letters refer to either the corresponding city districts or prominent features such as lakes or urban parks. The chosen locations cover a variety of geographic conditions, including 29 stations in built-up locations (Local Climate Zones \citep[LCZs; ][]{stewart2012local} 2, 5, 6, and 8), two stations in urban parks (FRSEEP (LCZ B) and FRFRIE (LCZ 9)), and eleven stations located outside of the built-up parts of the city. Among the latter were two stations at higher elevation (FRWITT (LCZ 9) at 444 meters above sea level (m a.s.l.) and FREICH (LCZ A) at 695 m a.s.l.), and a station in direct proximity of a lake (FROPFS (LCZ G)). More information on each station is given in Table \ref{tab:stations_overview}, and a detailed station documentation is provided by \cite{zenodo_station_docu}. The Freiburg WSN data are automatically transmitted and integrated into an urban near-time data management system \citep{zeeman2024modular}.

The raw data, with a temporal resolution of 30 seconds for the T1 stations and 1 minute for the T2 stations, were subjected to a quality control (QC) procedure, which consisted of three independent tests: A range test with limits of -35~°C and 45~°C ($T_a$) and 10\% and 100\% ($RH$), a rate of change test which allowed a maximum variation of 5~°C, 10~°C, and 15~°C or 10\%, 30\%, and 50\% over 1, 10, and 60 minutes, respectively, and a persistence check that flagged constant values occurring over periods of more than 6 hours ($T_a$) and 72 hours ($RH$). Any flagged data points were deleted. Additionally, the individual time series underwent a manual inspection, which resulted in the deletion of $RH$ measurements for stations FREICH from September 3 to 22 and from October 10 to November 28, 2022, and for station FRLORE from December 26, 2022, to May 25, 2023, due to recurring implausibly low measurements.

After applying the QC procedure, the total data availability was 95.7\% (first year) and 97.5\% (second year) for $T_a$ and 94.1\% (97.3\%) for $RH$. The slightly lower percentage of available data for the first year results mainly from the fact that a) three stations (FREBNE, FRMERZ, and FRTECH) were only installed in November 2023, b) three stations (FRASHA, FRSEEP, and FRVAUB) were non-functional for extended periods of the first year, and c) periods of $RH$ measurements from stations FREICH and FRLORE were flagged and removed from analysis upon manual investigation, as explained above. An overview of the station-wise data availability for both variables is given in Fig. \ref{fig:data_avail}.

To reduce the dependency of humidity on $T_a$, $RH$ was transformed to $e$ before modeling, using Teten's equation \citep{stull2015practical}: $e = 0.61078\ \text{exp}\left(\frac{17.7T_a}{T_a+237.3} \right)$. Moreover, $T_a$ and $e$ were averaged from a temporal resolution of 1 min to a resolution of 10 min and scaled to the interval $[0;1]$ to prevent the varying dimensions and units of measurement of the variables from affecting model training.

\subsection{Modeling}

\subsubsection{General modeling characteristics}

The modeling procedure was divided into three steps: (1) hyperparameter tuning to identify the optimal EGB  parameters for different numbers of predictor stations; (2) step-wise station removal; and (3) final model training, where the previously identified station subsets were used to train models. All of these modeling steps only use data from the first study year to prevent information leakage from the second year, which serves as independent evaluation data. Moreover, to obtain more robust and reliable results, all modeling steps were performed using 10-fold cross-validation (CV), where the daily mean $T_a$ across all stations was used to divide the data of the first year into 10 folds, each of which contained a similar distribution of days across the present range of $T_a$. In each model, eight folds were used for training (train folds), one for validation (val fold) to track training improvement and prevent overfitting, and one for evaluating the trained model (test fold). This means that each day of the study period was used once for model evaluation during the cross-validation procedure.

Training models that can predict all stations and both variables simultaneously required structuring the data such that each station's data could be used both as predictors and targets, while preventing the model from seeing the very same data it was supposed to estimate. To achieve this, the data were first transformed into a wide format, where each row corresponds to a timestamp and each column represents a variable at one specific station. Then, for each timestamp, $n$ stations were randomly selected as target stations, where $n$ is a random whole number between 1 and $N/4$, with $N$ the number of stations for which data was available at this timestamp. Then, a separate training instance was created for the $T_a$ and $e$ measurements of each selected station of that timestamp, while creating a target value column that holds the corresponding value and additional columns that specify the target variable and station. To prevent the target value from being included in the predictors, the $T_a$ and $e$ data of all stations selected as target stations for that specific time stamp were deleted from the set of predictors (set to Not Available (NA)). This procedure was executed for each timestamp in the training and validation data sets, whereas for the test data, each station was predicted at every timestamp for which it contained observations, meaning that the performance of the trained model was evaluated on all of the available data for each station.

Lastly, external predictors were added, which encompass both spatially invariant, temporal predictors (time of day and day of year) and temporally invariant, station-wise predictors (Sky View Factor (determined using Fisheye photographs and the software Rayman Pro \citep{matzarakis2007modelling, matzarakis2010modelling}) and elevation (derived from \cite{lgl})).

\subsubsection{Hyperparameter tuning}

Hyperparameter tuning was performed using a grid-based search, testing all combinations of the parameters \textit{learning rate} (values: 0.1, 0.3, and 0.5), \textit{tree depth} (values: 6 and 10), and \textit{training instance subsampling} (values: 0.5 and 1.0). The number of subsequent trees, itself a hyperparameter, was determined by monitoring model performance on the validation fold throughout training, which was stopped once performance did not improve over 500 trees. Since the modeling procedure involved training models with a varying number of predictor stations, which can affect the choice of hyperparameters, we performed hyperparameter optimization separately for each subset size. Moreover, hyperparameter tuning was performed in the previously described CV setting. Concretely, for each test fold and station subset size, a randomly selected subset of stations was chosen to train a model with the respective hyperparameter combination. The model's performance was then evaluated on the respective test fold using root-mean-square error (RMSE). The hyperparameter combination that achieved the lowest average RMSE across all folds was then used to train all models with the corresponding number of predictor stations in subsequent modeling steps.

\subsubsection{Station removal}
To determine the optimal order for station removal while maintaining the best possible model performance across all stations, we developed a backward elimination procedure, which is illustrated in Algorithm~\ref{alg:station_selection}. Starting with a full model trained using all 42 stations as predictors, the algorithm iteratively evaluated the importance of each predictor station by temporarily setting its data to NA and calculating the resulting increase in prediction RMSE across all stations and both variables on the test fold. The station whose removal yielded the smallest RMSE increase was then permanently set to NA before the procedure was repeated with the remaining stations. To prevent the accumulating NA values from distorting model training as more stations were removed, the model was retrained at predefined points (when the remaining station count reached 35, 28, 21, 14, 10, 7, 4, 3, or 2 remaining stations) using only the currently remaining predictors. The retraining steps were chosen to balance model accuracy with computational efficiency, as, on the one hand, retraining at every elimination step would be computationally prohibitive given the total number of required models and, on the other hand, setting too many stations' data to NA between retraining might have deteriorated model performance and obfuscated the patterns the model was supposed to learn. The entire station removal procedure was executed independently for each of the 10 cross-validation test folds, yielding 10 distinct station removal sequences from which optimal subsets were extracted for final model training.

\begin{algorithm}[ht!]
\caption{Pseudo-code of station removal procedure.}
\label{alg:station_selection}
\begin{algorithmic}[1]
\STATE \textbf{Initialize:}
\STATE \textit{retraining\_points} $\gets \{35, 28, 21, 14, 10, 7, 4, 3, 2\}$
\STATE \textit{removed\_stations} $\gets \emptyset$
\STATE \textit{removed\_since\_retrain} $\gets \emptyset$
\STATE \textit{current\_stations} $\gets \{1, 2, \ldots, 42\}$
\STATE \textit{current\_model} $\gets$ Train full model using all 42 stations
\STATE
\FOR{$n = 42$ \textbf{down to} $2$}
    \IF{$n \in \textit{retraining\_points}$}
        \STATE \textit{current\_model} $\gets$ Retrain model excluding all stations in \textit{removed\_stations}
        \STATE \textit{removed\_since\_retrain} $\gets \emptyset$
    \ELSE
        \STATE Set predictors $T_a$ and $e$ of all stations in \textit{removed\_since\_retrain} to NA 
    \ENDIF
    \STATE
    \STATE \textit{best\_rmse} $\gets \infty$
    \STATE \textit{station\_to\_remove} $\gets$ null
    \STATE
    \FORALL{$s \in \textit{current\_stations}$}
        \STATE Set predictors $T_a$ and $e$ of station $s$ to NA
        \STATE Predict on test set using \textit{current\_model} with remaining stations
        \STATE \textit{rmse} $\gets$ Calculate RMSE averaged across all stations and both variables
        \STATE
        \IF{\textit{rmse} $<$ \textit{best\_rmse}}
            \STATE \textit{best\_rmse} $\gets$ \textit{rmse}
            \STATE \textit{station\_to\_remove} $\gets$ s
        \ENDIF
        \STATE
        \STATE Restore predictors $T_a$ and $e$ of station $s$
    \ENDFOR
    \STATE
    \STATE \textit{current\_stations} $\gets$ \textit{current\_stations} $\setminus$ \{\textit{station\_to\_remove}\}
    \STATE \textit{removed\_stations} $\gets$ \textit{removed\_stations} $\cup$ \{\textit{station\_to\_remove}\}
    \STATE \textit{removed\_since\_retrain} $\gets$ \textit{removed\_since\_retrain} $\cup$ \{\textit{station\_to\_remove}\}
\ENDFOR
\end{algorithmic}
\end{algorithm}

\subsubsection{Fitting final models}
To train the final models, station subsets were extracted from the elimination sequences obtained in the previous step. To avoid overly optimistic performance estimates resulting from potential overfitting to the test folds used during station removal, new random CV folds were generated for final model training. Consistent with the station selection procedure, only data from the first study year were used for model training, while performance evaluation was conducted on the independent dataset from the second year. Therefore, the approach simulated a realistic scenario where station removal decisions must be made after the first year of WSN operation, with model performance subsequently assessed on unseen data from the following year.

\subsubsection{Reference models}
In addition to the previously introduced models, which were trained on the first study year and evaluated on the second (hereafter referred to as {EGB  1$\rightarrow$2}), several reference and baseline models were developed to evaluate different aspects of model performance and the station selection procedure. These include three EGB  variants, as well as one simple regression model and one numerical urban climate model. 

\paragraph{EGB variants}

{EGB  1$\rightarrow$1} refers to the same model approach just outlined ({EGB  1$\rightarrow$2}) but was evaluated on the corresponding test folds of the \textit{first} study year (the training year), allowing for an assessment of temporal extrapolation effects compared to predictions within the training period. {EGB  1,2$\rightarrow$1,2} employed the same predictor station subsets as the main models but were trained and evaluated on data from \textit{both} study years, enabling evaluation of whether one year of WSN operation captures all relevant patterns or if longer operational periods provide additional benefits. Additionally, all model variants ({EGB  1$\rightarrow$1}, {EGB  1$\rightarrow$2}, and {EGB  1,2$\rightarrow$1,2}) were retrained using randomly selected predictor station subsets instead of those resulting from the station removal procedure, allowing assessment of the benefits provided by the systematic station selection approach over random station subsets. For robust performance estimates, ten random predictor station subsets were generated for each subset size and test fold.

\paragraph{Statistical and numerical models}

The statistical baseline models were GLMs trained separately for each target station and variable. These models utilize data from two official weather stations operated by the DWD as predictors: station 01443, located northwest of the city on the premises of the local airport, and station 13667, situated in the city center (Fig. \ref{fig:study_area}). The GLMs follow the functional form:
\begin{equation}
\hat{y}_{i,v} = \beta_0 + \beta_1 X_{1,v} + \beta_2 X_{2,v} + \beta_3 (X_{1,v} \times X_{2,v})
\end{equation}
where $\hat{y}_{i,v}$
represents the estimated value of variable $v$
($T_a$ or $e$) at WSN station $i$; $X_{1,v}$ and $X_{2,v}$ denote the corresponding variable's measurements from DWD stations 01443 and 13667, respectively, and $\beta_0$, $\beta_1$, $\beta_2$, and $\beta_3$ are the model coefficients representing the intercept, linear effects of both official stations, and their interaction term. The GLM represents a comparatively simple baseline model that indicates the predictive accuracy achievable with just two reference stations (one urban and one rural station), linearly modeled relationships, and no additional predictors.

Lastly, {SUEWS} represents a state-of-the-art, neighborhood-scale urban land surface model \citep{jarvi2011surface, ward2016surface, sun2019python}. It simulates urban energy, water, and mass fluxes and has been evaluated in multiple cities and climates, including Freiburg \citep{briegel2024high}. In the offline version deployed for this study, it uses meteorological forcing data from a weather station operated by the University of Freiburg, located on a rooftop at approximately 55 m a.g.l. and in close proximity to the station FRINST (48.0011°N, 7.8486°E, see Fig. \ref{fig:study_area}), in addition to local-scale aggregated spatial and socio-economic data. {SUEWS} was used to provide hourly estimates of $T_a$ and $RH$ at 2 m a.g.l. for grid cells of 500 m $\times$ 500 m. These estimates were evaluated against the WSN stations located in the respective cells (note that the WSN stations were mounted at 3 m a.g.l.). Due to its restricted spatial modeling domain, {SUEWS} did not produce estimates for the stations FREBNE, FREICH, FRMERZ, FRWAHS, FRWILD, FRWITT, and FRZAHR, which were therefore excluded from the comparisons between \textit{SUEWS} and the EGB  models. This study used the Python implementation of SUEWS (supy version 2022.9.22; supy-driver version 2021a5). Further details regarding its requirements and implementation are provided by \cite{briegel2024high}. 

Since all models used in this study only produce one humidity measure ($e$ in the case of the EGB  models and $RH$ for GLM and {SUEWS}), the respective missing variable was derived using Teten's formula to enable comparisons across all three variables.

\section{Results}
\label{Results}

\subsection{Hyperparameter search}

For all numbers of predictor stations, the hyperparameter combinations featuring maximum tree depth (10), a learning rate of 0.1, and no subsampling of training instances led to the smallest RMSE during model training (Fig.~\ref{fig:hyp_opt}). This indicates the presence of complex and non-linear interactions among predictors, which require deep individual trees to capture intricate patterns and moderate learning rates to ensure gradual, stable convergence during the iterative boosting process.

\subsection{Imputation accuracy}

Overall, {EGB  1$\rightarrow$2} models achieved a good imputation accuracy across the range of used predictor stations and for all variables, with RMSEs from 0.69 K to 0.93 K for $T_a$, 3.84\% to 4.88\% for $RH$ and 0.72 hPa to 0.9 hPa for $e$ when comparing the full models with 42 stations to those using only 2 stations (Table~\ref{tab:error_metrics}). For all variables, the decrease in predictive accuracy is more pronounced for lower numbers of predictor stations, while WSN density reductions of up to 66\%, corresponding to 14 remaining stations, only have minor effects on prediction accuracy. Notably, mean absolute errors (MAEs) were consistently around 25-35\% lower than the corresponding RMSE values across all variables, indicating the presence of occasional large deviations between model predictions and observations. The coefficient of determination (R\textsuperscript{2}) scores were higher for $T_a$ (0.992-0.984) than for both humidity measures (0.955-0.931 for $RH$ and 0.983-0.973 for $e$). Mean bias errors (MBEs) were consistently negative and increased slightly with higher degrees of WSN density reduction, but were overall small in magnitude. Note, however, that the MBEs displayed here only allow for limited interpretation as the values were averaged across stations. The fact that the smallest MBE for $RH$ with -0.001 was achieved by models using only four predictor stations, for instance, suggests the presence of larger positive and negative biases among individual stations that effectively cancel each other. Therefore, a detailed analysis of prediction biases is provided in a later section.

\begin{table}[ht!]
\centering
\caption{Error metrics for {EGB  1$\rightarrow$2} models for all variables and numbers of predictor stations, averaged across target stations. RMSE = root-mean-square Error, MAE = mean absolute error, R\textsuperscript{2} = coefficient of determination, MBE = mean bias error.}
\label{tab:error_metrics}
\resizebox{\textwidth}{!}{%
\begin{tabular}{@{}cccccccccccc@{}}
\toprule
Variable            & Error metric & \multicolumn{10}{c}{Number of predictor stations retained}                                       \\
                    &              & 42    & 35    & 28    & 21    & 14    & 10    & 7     & 4     & 3     & 2     \\ \midrule
\multirow{4}{*}{$T_a$} & RMSE         & 0.69  & 0.70  & 0.71  & 0.71  & 0.73  & 0.75  & 0.78  & 0.83  & 0.87  & 0.93  \\
                    & MAE          & 0.49  & 0.50  & 0.50  & 0.51  & 0.53  & 0.54  & 0.56  & 0.60  & 0.63  & 0.68  \\
                    & R\textsuperscript{2}           & 0.992 & 0.991 & 0.991 & 0.991 & 0.990 & 0.990 & 0.989 & 0.988 & 0.987 & 0.984 \\
                    & MBE          & -0.019 & -0.030 & -0.026 & -0.027 & -0.024 & -0.023 & -0.015 & -0.019 & -0.030 & -0.041 \\
                    \midrule
\multirow{4}{*}{$RH$} & RMSE         & 3.84  & 3.87  & 3.92  & 3.96  & 4.05  & 4.13  & 4.21  & 4.44  & 4.63  & 4.88  \\
                    & MAE          & 2.95  & 2.96  & 2.99  & 3.02  & 3.09  & 3.15  & 3.21  & 3.41  & 3.54  & 3.72  \\
                    & R\textsuperscript{2}           & 0.955 & 0.954 & 0.953 & 0.953 & 0.951 & 0.949 & 0.947 & 0.942 & 0.937 & 0.931 \\
                    & MBE          & -0.197 & -0.227 & -0.297 & -0.321 & -0.325 & -0.424 & -0.345 & -0.001 & -0.305 & -0.500 \\
                    \midrule
\multirow{4}{*}{$e$} & RMSE         & 0.72  & 0.72  & 0.73  & 0.74  & 0.76  & 0.77  & 0.79  & 0.83  & 0.87  & 0.90  \\
                    & MAE          & 0.50  & 0.50  & 0.50  & 0.51  & 0.52  & 0.53  & 0.54  & 0.58  & 0.60  & 0.63  \\
                    & R\textsuperscript{2}           & 0.983 & 0.983 & 0.982 & 0.982 & 0.981 & 0.980 & 0.979 & 0.977 & 0.975 & 0.973 \\
                    & MBE          & -0.072 & -0.076 & -0.091 & -0.097 & -0.101 & -0.125 & -0.109 & -0.057 & -0.119 & -0.160 \\ \bottomrule 
\end{tabular}
}
\end{table}

\begin{figure}[ht!]
 \centerline{\includegraphics[width=39pc]{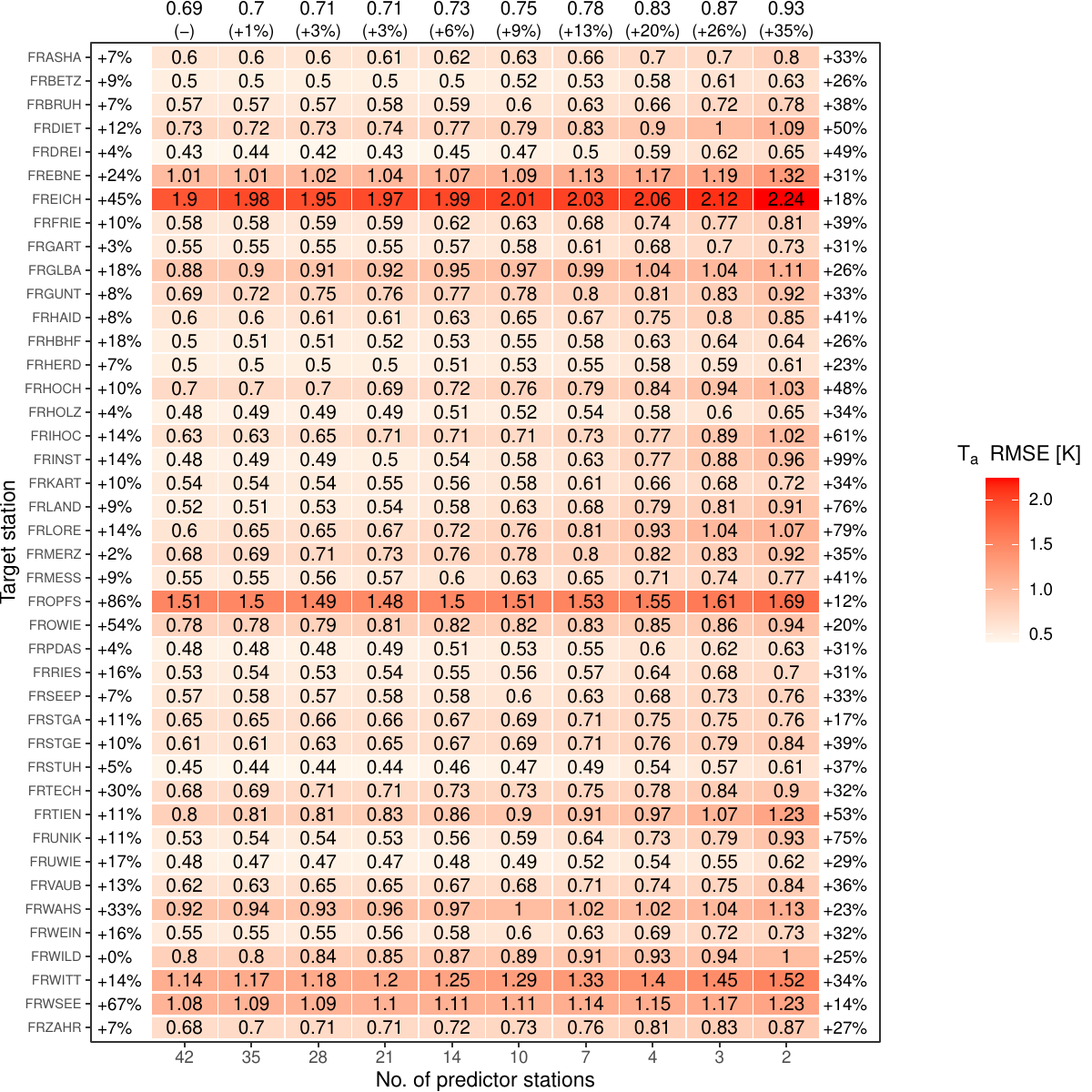}}
  \caption{Station-wise $T_a$ prediction RMSEs of {EGB  1$\rightarrow$2} models across the number of stations retained. Values on top of the plot denote the average RMSE across all target stations and, in parenthesis, the percentage increase averaged across all stations compared to the full models, percentages on the left refer to the increase in RMSE relative to the {EGB  1$\rightarrow$1} models, percentages on the right give the station-wise percentage RMSE increase of the models featuring two stations relative to the full models.}
  \label{fig:rmse_no_stations_ta}
\end{figure}

\begin{figure}[ht!]
 \centerline{\includegraphics[width=39pc]{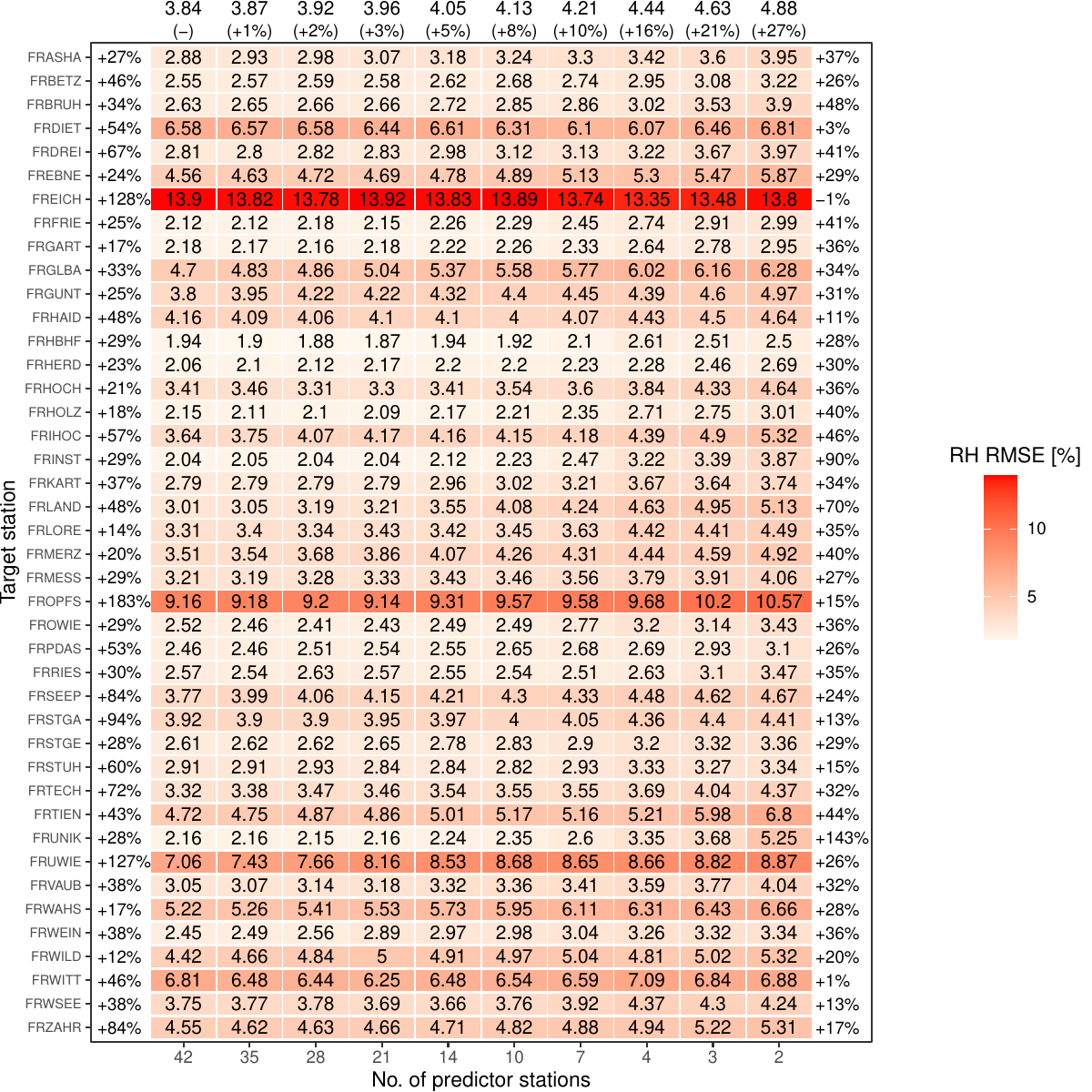}}
  \caption{Station-wise $RH$ prediction RMSEs of {EGB  1$\rightarrow$2} models across the number of stations retained. See Fig. \ref{fig:rmse_no_stations_ta} for further explanations.}
  \label{fig:rmse_no_stations_rh}
\end{figure}

A station-wise investigation of prediction errors over the range of used predictor stations revealed interesting patterns for both $T_a$ (Fig. \ref{fig:rmse_no_stations_ta}) and $RH$ (Fig. \ref{fig:rmse_no_stations_rh}; the corresponding figure for $e$ and figures depicting the RMSEs for the first study year ({EGB  1$\rightarrow$1}) are provided in the Appendix Figs. \ref{fig:rmse_no_stations_vp}, \ref{fig:rmse_no_stations_ta_first_year}, \ref{fig:rmse_no_stations_rh_first_year}, and \ref{fig:rmse_no_stations_vp_first_year}). 
First, RMSEs varied greatly between target stations, with larger errors at remote stations compared to those in built-up, central locations. The highest RMSEs of full models for both $T_a$ and $RH$ were found at the forest stations FREICH (1.9 K and 13.9\%) and FROPFS (1.51 K and 9.16\%), and the smallest errors were recorded at the central stations FRDREI for $T_a$ (0.43 K) and FRHBHF for $RH$ (1.94\%).

Second, the deterioration of model performance when fewer predictor stations were used differs between stations (right-most columns). Stations with higher errors for the full model generally exhibit a low percentage increase in RMSE and, in the case of stations FREICH and FRWITT for $RH$, errors even remain virtually unchanged (-1\% and +1\% respectively). For stations showing low RMSEs in the full models, the pattern was more varied: In some cases (FRHBHF and FRHERD for $T_a$ and FRHBHF and FRSTUH for $RH$), the increase was comparatively low (15-28\%), whereas FRUNIK and FRINST experienced a high deterioration of prediction accuracy, with increases of 75\% and 99\% for $T_a$ and 145\% and 90\% for $RH$. 

Third, the deterioration of the models compared to the {EGB  1$\rightarrow$1} models (left-most column), which can be interpreted as the models' ability to extrapolate temporally, showed lower values for $T_a$ compared to $RH$ for most stations, except for FROWIE, FRWAHS, and FRWSEE. The stations that exhibited high overall RMSEs generally exhibited higher percentage increases compared to the {EGB  1$\rightarrow$1} models, although there are exceptions to this pattern, such as for stations FRWAHS and FRWITT. Especially with regards to $T_a$, there are several stations whose errors show almost no differences between the study years, including FRWILD, with an increase of 0\%, FRMERZ, with 3\%, and FRDREI, with 4\%. Finally, FRUWIE and FRSTGA represent interesting cases in that their respective deteriorations during the second study year are remarkably lower for $T_a$ (17\% and 11\%) compared to $RH$ (127\% and 94\%). 

\begin{figure}[ht!]
 \centerline{\includegraphics[width=39pc]{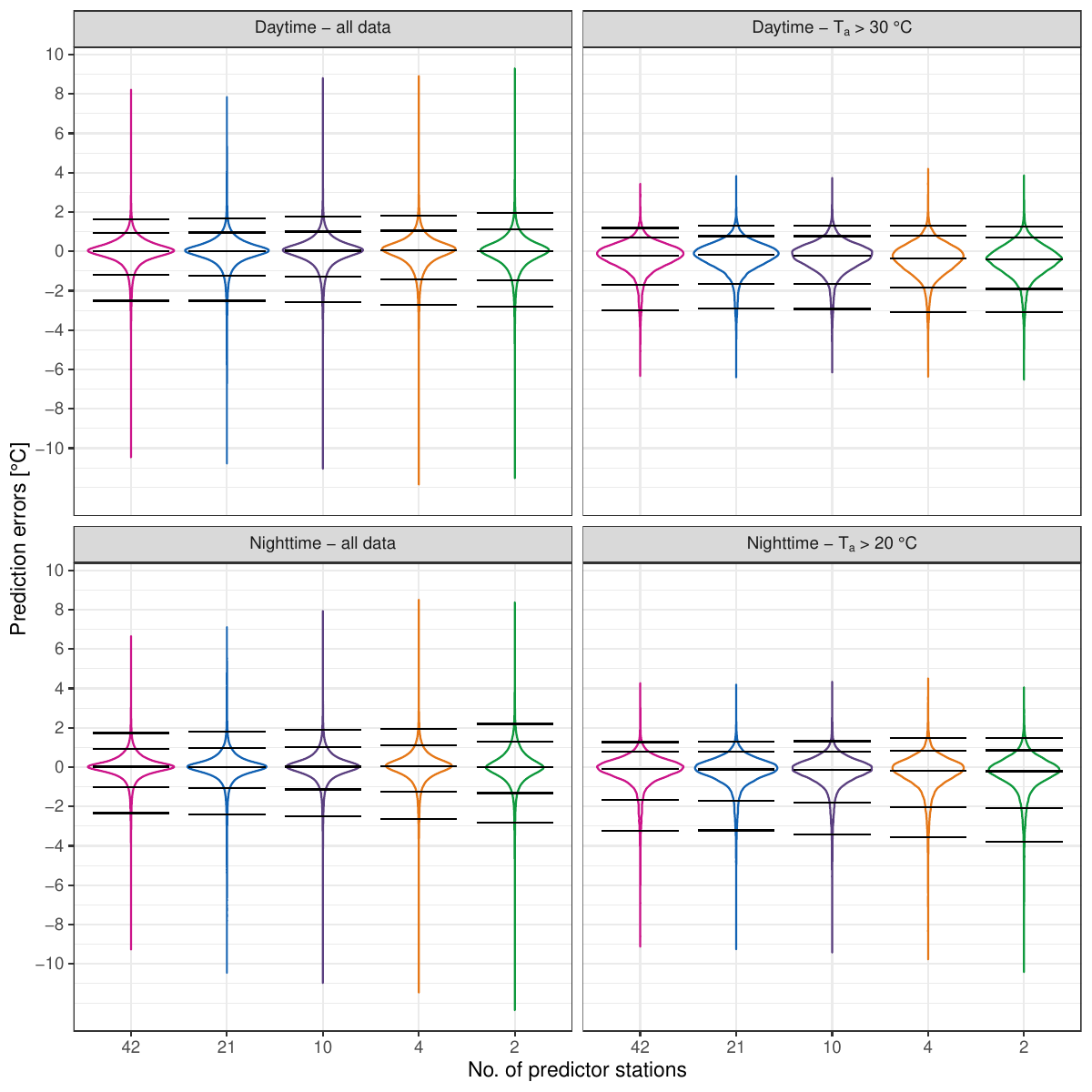}}
  \caption{$T_a$ prediction errors of {EGB  1$\rightarrow$2} models across all stations and the entire second study year, split by daytime (top) vs. nighttime (bottom) and for the entire study period (left) vs. under hot conditions (right). Black horizontal lines denote the 1st, 5th, 50th (median), 95th, and 99th error percentiles.}
  \label{fig:boxplot_hot_periods}
\end{figure}

Analyzing prediction errors separately for day- and nighttime and for observations exhibiting high values of $T_a$ enables the evaluation of model performance in different meteorological situations (Fig. \ref{fig:boxplot_hot_periods}). Generally, the error distributions (indicated by the height of the violin plots and the vertical distribution of the percentiles drawn in black) appear similar for all plots within a given category, suggesting comparable prediction accuracy across the number of predictor stations, regardless of the time and $T_a$ conditions. The median errors, indicative of a systematic prediction bias, show no visible deviation from zero for the entire set of observations, both during the day and at nighttime. In contrast, for the hot observations, small negative deviations can be identified, particularly for the plots corresponding to lower numbers of predictor stations. Across all data categories, the 90th and the 98th percentile ranges show that model deviations rarely exceeded 2 K in either direction, but both percentiles display a slight negative bias, suggesting that extreme negative prediction errors were more common than their positive counterparts, which is also indicated by the larger maximum negative errors visible. The rare extreme errors (exceeding 8 K in either direction) were found to occur exclusively at stations at the geographic edges of the study area (FREBNE, FREICH, FRHOCH, FRIHOC, FRLAND, FROPFS, FRTIEN, FRWILD, FRWITT, and FRWSEE).

Comparing the predicted frequency of various climatic indicator days predicted by {EGB  1$\rightarrow$2} models to those observed during the second study year allows for an assessment of the ability of thinned-out WSNs to reproduce accurate yearly climate statistics across stations over periods following the station removal (Table \ref{tab:clim_ind_errors}). While all {EGB  1$\rightarrow$2} models consistently underestimated the occurrence of each type of indicator days, the extent of misrepresentation differed both between the type of indicator day and the number of stations retained, that is, the degree to which the WSN was thinned. For summer days, hot days, and frost days, all featuring higher values than the other climatic indicators, the deviations between estimates and observations, both absolute and in percentages, were generally small (between -2.3\% and -9.3\%) and became larger as fewer predictor stations were used. For the desert days, ice days, and tropical nights, the absolute deviations were still small, with a maximum of -0.4 days for the desert and ice days, and -1.2 days for the tropical nights, but due to the lower observed frequency of these indicators, the percentage deviations were larger. Interestingly, the worst estimates for both ice days and tropical nights were produced by models using 10 predictor stations, with deviations of -7.8\% and -16.9\%, whereas the models using only 2 predictor stations achieved deviations of only -1.6\% and -14.5\%, respectively.

\begin{table}[]
\centering
\caption{Mean frequency of climatic indicator days for the second study year across all stations predicted by {EGB  1$\rightarrow$2} models with selected numbers of predictor stations compared to gap-filled data using the {EGB  1,2$\rightarrow$1,2} models (No. of predictor stations = 42). Percentages indicate the change relative to the gap-filled data. The definitions of each climatic indicator are provided in Table \ref{tab:clim_stats_crit}. For additional details regarding their calculation, see \cite{plein2025gapfilling}.}
\label{tab:clim_ind_errors}
\begin{tabular*}{\textwidth}{@{\extracolsep{\fill}}lccccc@{}}
\toprule
Climatic indicator     & \multicolumn{5}{c}{No. of predictor stations}                         \\
                       & 42    & 21            & 10            & 4             & 2             \\
                       \midrule
No. of summer days     & 101.1 & 98.8 (-2.3\%) & 98.4 (-2.7\%) & 97.6 (-3.5\%) & 96.8 (-4.3\%) \\
No. of hot days        & 32.3  & 31 (-4\%)     & 30.7 (-5\%)   & 30.2 (-6.5\%) & 29.3 (-9.3\%) \\
No. of desert days     & 2.1   & 2 (-4.8\%)    & 1.9 (-9.5\%)  & 1.7 (-19\%)   & 1.7 (-19\%)   \\
No. of frost days      & 25.8  & 24.8 (-3.9\%) & 24.3 (-5.8\%) & 24.1 (-6.6\%) & 23.4 (-9.3\%) \\
No. of ice days        & 6.4   & 6 (-6.2\%)    & 5.9 (-7.8\%)  & 6.3 (-1.6\%)  & 6.3 (-1.6\%)  \\
No. of tropical nights & 8.3   & 7.4 (-10.8\%) & 6.9 (-16.9\%) & 7 (-15.7\%)   & 7.1 (-14.5\%) \\ \bottomrule
\end{tabular*}
\end{table}

\begin{figure}[ht!]
 \centerline{\includegraphics[width=33pc]{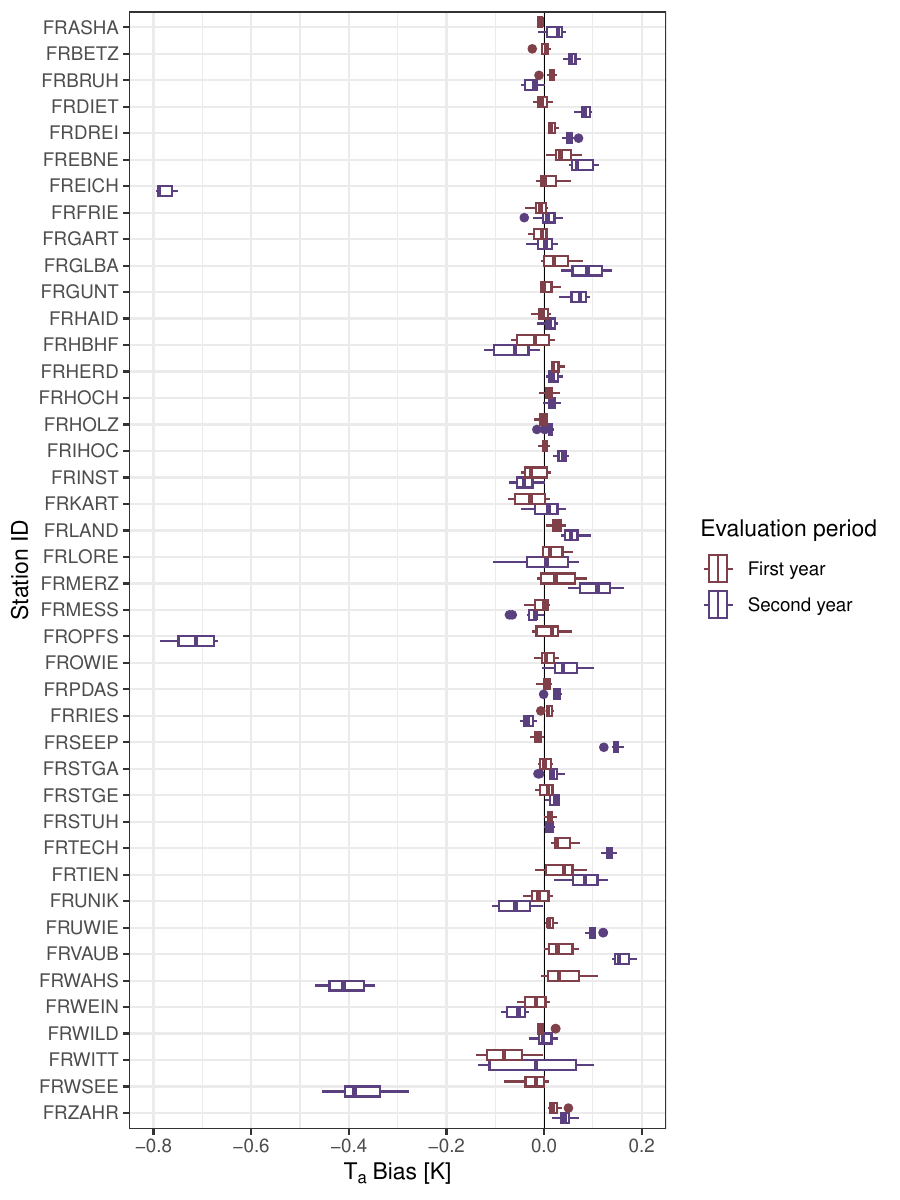}}
  \caption{Station-wise prediction bias of $T_a$ across CV folds for the first ({EGB  1$\rightarrow$1}) and second ({EGB  1$\rightarrow$2}) study year.}
  \label{fig:boxplots_bias_ta}
\end{figure}

A station-wise comparison of prediction biases for $T_a$ between the first and second study year shows large differences between stations (Fig. \ref{fig:boxplots_bias_ta}; station-wise mean biases for all variables are provided in Table \ref{tab:bias_all_vars}). The majority of stations display small biases for both study years (-0.2 to 0.2 K), with only slightly larger values for the second study year. There are, however, several exceptions, most notably for the forest locations FREICH, FROPFS, FRWAHS, as well as for FRWSEE, all with large negative biases for the second year, with small to no biases during the training period, and the suburban station FRVAUB, with a smaller, but consistent positive bias for the second year. The largest bias during the second study year is found at FREICH with -0.79 K, followed by FROPFS with -0.72 K. Notably, remote stations located near or outside of forests (e.g., FRTIEN, FREBNE, FRWILD, FRWITT) do not exhibit large biases. Moreover, the large bias found at station FRWSEE is not present in the geographically adjacent stations FRASHA and FREBNE. 

\begin{figure}[ht!]
 \centerline{\includegraphics[width=33pc]{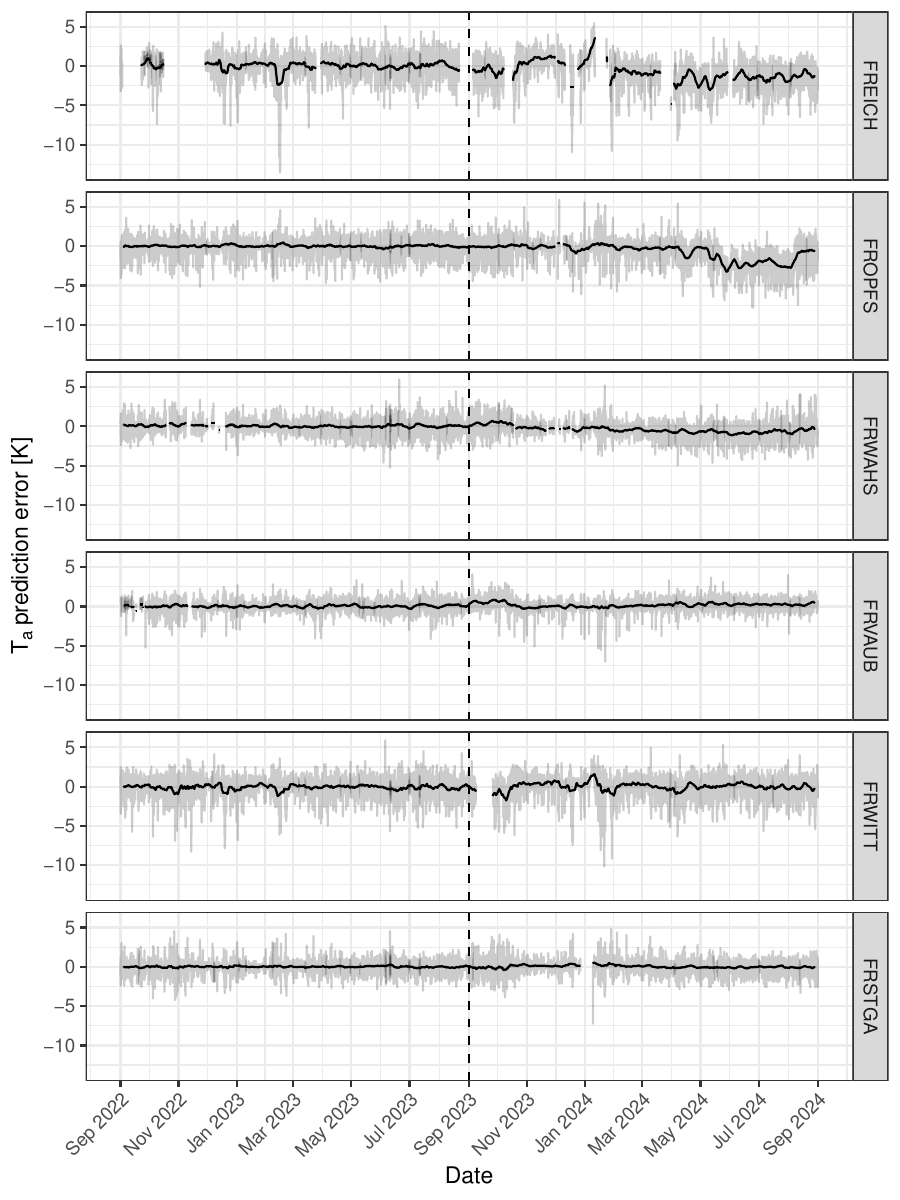}}
  \caption{$T_a$ model prediction errors (grey) and moving 7-day average errors (black) for the study period and selected stations. The vertical dashed line separates the first (training) from the second (evaluation) study year.}
  \label{fig:timeseries_bias_ta}
\end{figure}

The temporal progression of prediction errors throughout the entire study period is displayed for selected stations in Fig. \ref{fig:timeseries_bias_ta}. The top four plots show stations exhibiting large biases, while the bottom two represent stations with small biases: one station with large prediction errors (FRWITT) and one with small prediction errors (FRSTGA). During the first study year, the moving average indicates no consistent bias over extended periods for any of the displayed stations, even though prediction errors differed substantially. Over the course of the second year, the patterns varied between stations: In addition to large gaps resulting from poor reception, vandalism, and sensor malfunctioning, FREICH displayed large fluctuations in the 7-day moving average prediction errors, with a period of positive biases during January 2024, followed by consistently but varyingly pronounced negative biases throughout the remaining study period. FROPFS showed no prolonged periods of high bias until April 2024, after which pronounced and consistent negative biases are visible. FRWAHS began showing negative biases earlier in March 2025, though at a much smaller magnitude compared to FREICH and FROPFS. FRVAUB, unlike the other stations, displayed a positive bias, which is clearly visible during the first 1.5 months of the second study year, and more subtle thereafter. At FRWITT, two periods with consistent biases are visible, each spanning between one and two weeks: the first one, featuring negative biases, occured during a comparably warm period in the first half of October 2023 (see also Fig. \ref{fig:data_avail}), while the second one took place during the coldest period of the second study year in January 2024, and displayed positive biases. Apart from that, however, no consistent biases were present. FRSTGA, lastly, represents the patterns found at the majority of stations, with low overall prediction errors and no distinguishable biases during either of the study years.

\begin{figure}[ht!]
 \centerline{\includegraphics[width=27pc]{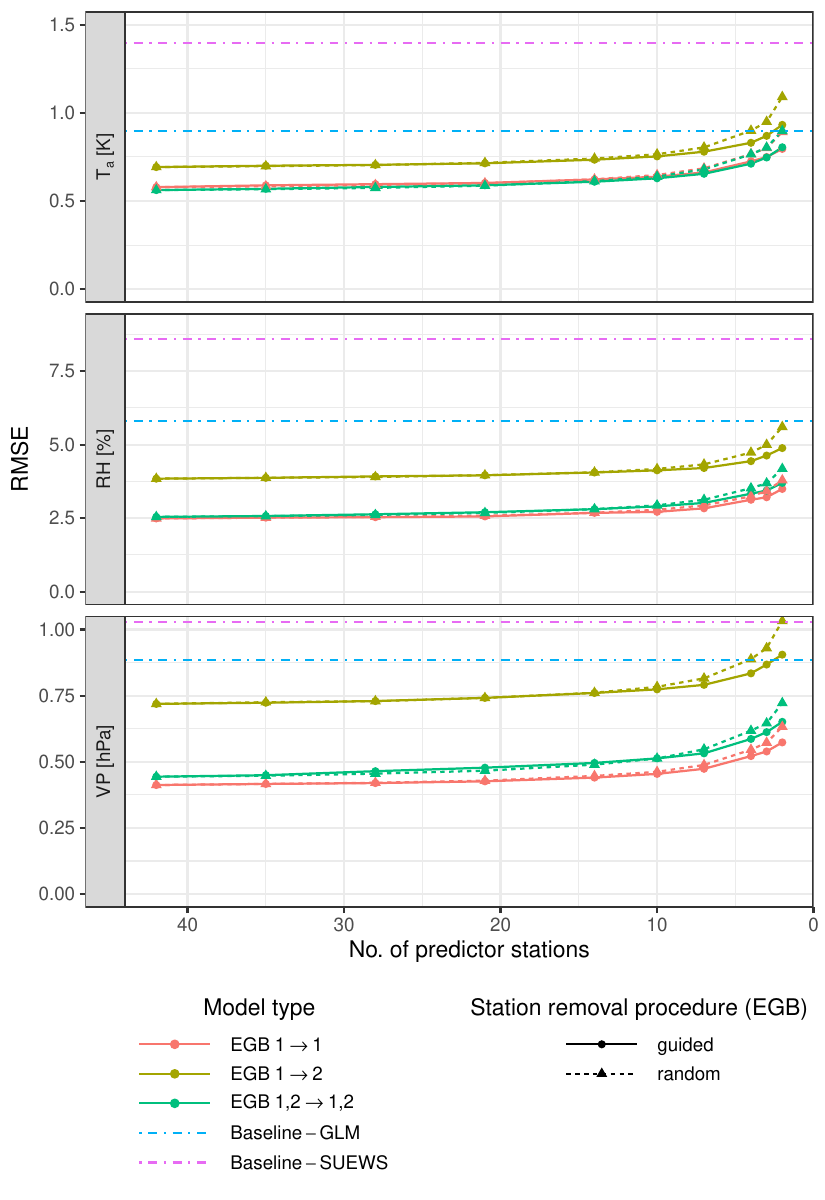}}
  \caption{Prediction RMSEs across the number of predictor stations for guided and random station removal procedures and reference models.}
  \label{fig:rmse_no_stations}
\end{figure}

Comparing predictive accuracy between model types reveals substantial differences (Fig. \ref{fig:rmse_no_stations}). Of all models, the numerical reference model {SUEWS} displayed the highest prediction errors across all variables, surpassing even the comparably simple statistical {GLM}s trained on two official DWD stations. This is particularly surprising given that some of the stations showing the largest prediction errors, particularly FREICH, FRWAHS, and FRWITT, were not even included in the averaged {SUEWS} errors. Both baseline models performed worse than all EGB  variants, apart from some {EGB  1$\rightarrow$2} variants in the lowest ranges of used predictor stations. Regarding the EGB models, the guided station removal procedure performed only marginally better than the random variant, even though the differences became slightly more pronounced as the number of predictor stations decreased, indicating that the choice of retained station subsets became more important for stronger degrees of WSN density reduction. The differences between the {EGB  1$\rightarrow$1} and {EGB  1,2$\rightarrow$1,2} models were generally small, suggesting that a second year of WSN operation provided little additional knowledge compared to using only one year for gap-filling (temporal interpolation). Temporal extrapolation, on the other hand, appeared to result in substantially higher prediction errors for all variables, with differences in RMSEs of approximately 0.1 K ($T_a$), 1.5\% ($RH$), and 0.3 hPa ($e$) between the {EGB  1$\rightarrow$2} models and the  {EGB  1$\rightarrow$1} and {EGB  1,2$\rightarrow$1,2} variants.

\clearpage

\subsection{Spatial configuration of station subsets}

\begin{figure}[ht!]
 \centerline{\includegraphics[width=39pc]{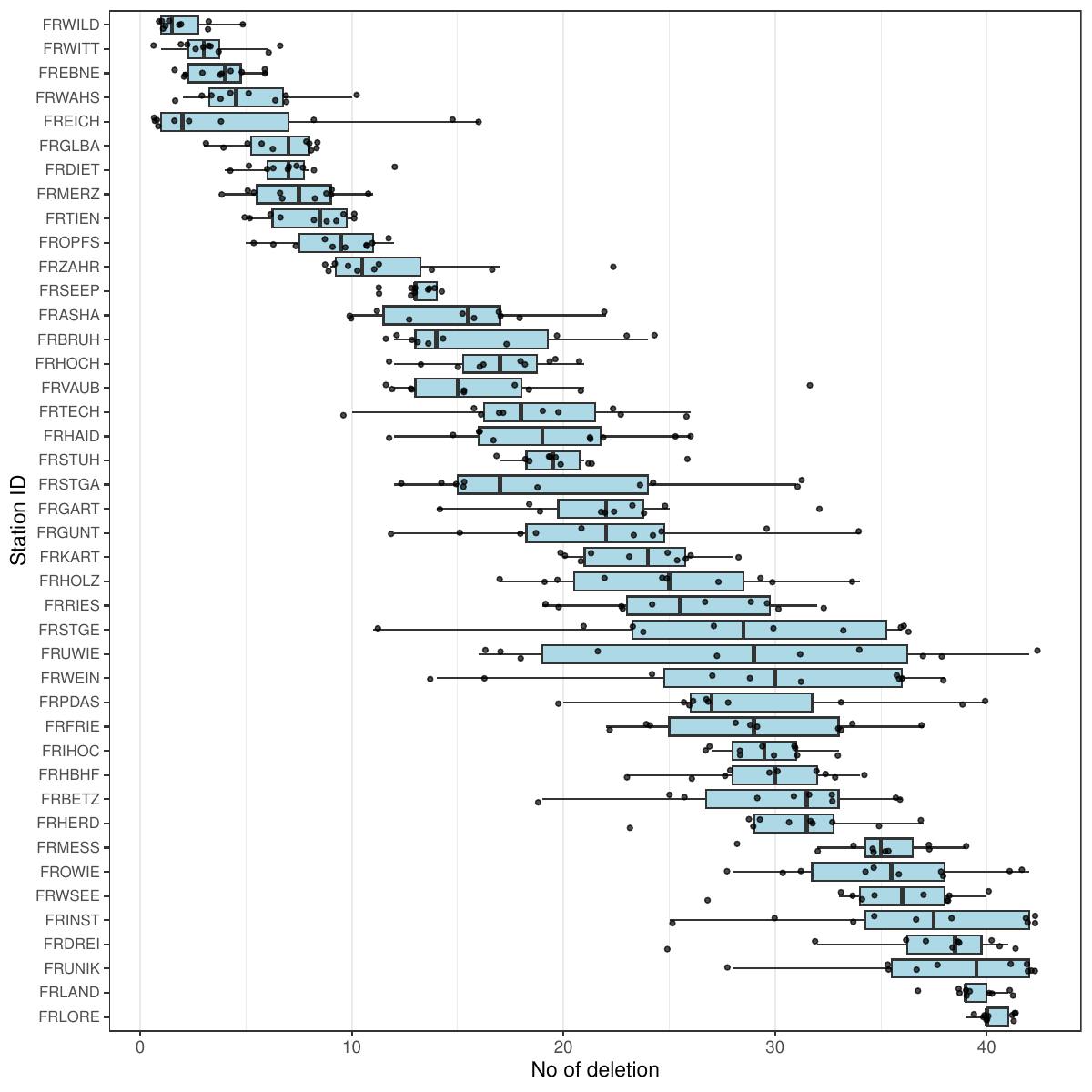}}
  \caption{Removal order of the WSN stations across CV folds, sorted in increasing order by the mean removal rank.}\label{fig:boxplots_del_order}
\end{figure}

\begin{figure}[ht!]
 \centerline{\includegraphics[width=27pc]{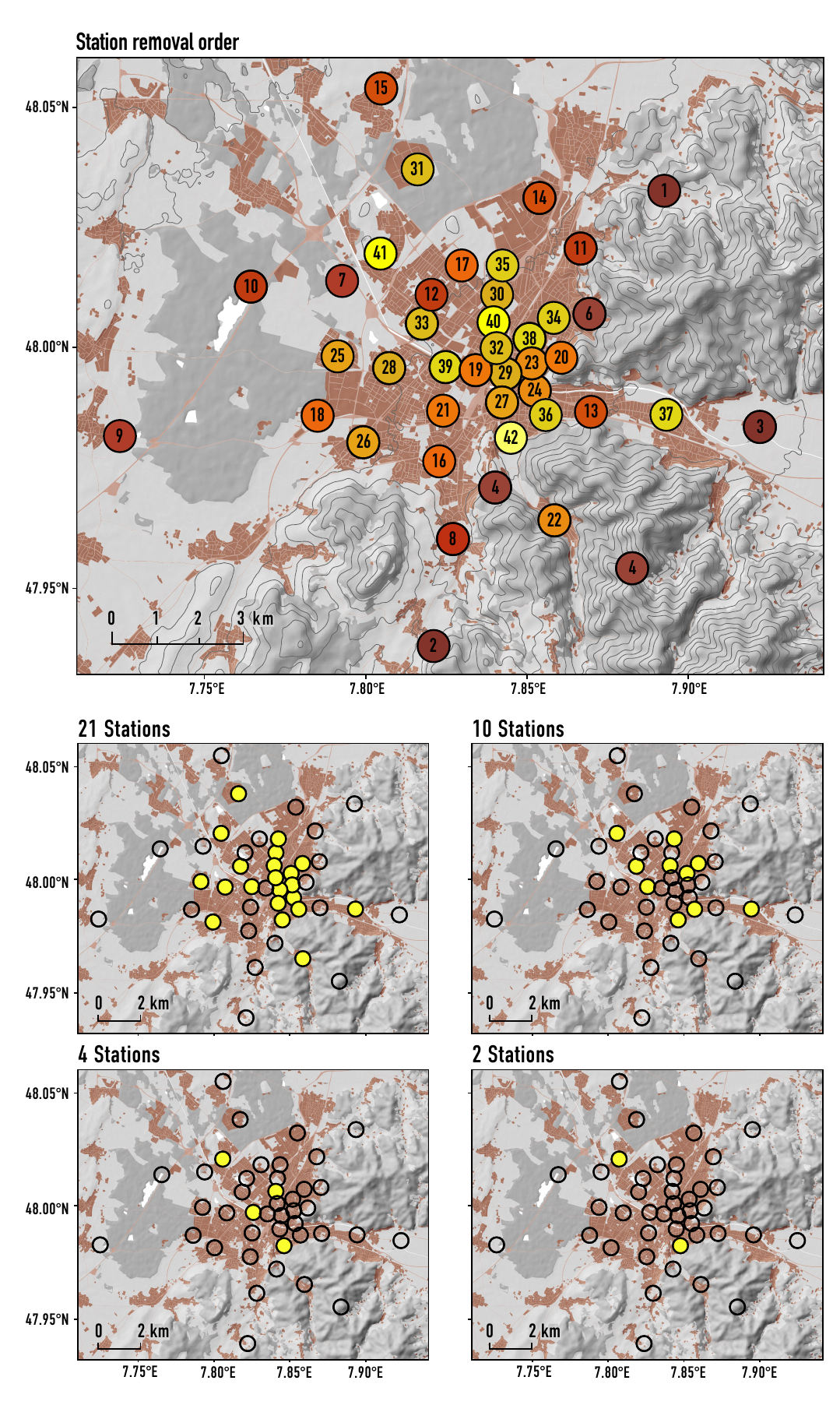}}
  \caption{Order of station removal (top) and remaining stations at selected subset sizes, averaged across CV folds. Note that rank 4 is shared by stations FRWAHS and FREICH.}
  \label{fig:stat_del_rank}
\end{figure}

Analyzing the order in which stations were removed from the WSN (Fig. \ref{fig:boxplots_del_order}) and the resulting spatial configurations of the thinned-out WSN (Fig. \ref{fig:stat_del_rank}) shows that, except for FROPFS, remote stations in non-built-up locations were consistently removed first, namely FRWILD, FRWITT, FREBNE,  FRWAHS, and FREICH. In contrast, stations located at the fringes of the WSN but within built-up areas, such as FRIHOC, FRASHA, FRRIES, and FRGUNT, were retained for longer, as were the majority of central stations, which is reflected in the subset containing 21 (50\%) of the original WSN. At 10 remaining stations (a reduction of approximately 76\%), more central stations were removed, along with all stations located in the southwestern parts of the city, even though there was considerable variability in the removal order of these stations between CV folds (Fig. \ref{fig:boxplots_del_order}). At four remaining stations, corresponding to a WSN reduction by approximately 90\%, two relatively adjacent central stations remained (FRDEI and FRUNIK), along with two stations located at the southeastern (FRLORE) and northwestern (FRLAND) edges of the main built-up city body. The latter two were the last remaining stations in a WSN reduced to only two stations and consistently retained the longest across CV folds. The apparent significance of the station FRLORE is surprising given that more than 40\% of its RH measurements during the first study year, which was used for the station removal procedure, were deleted due to sensor malfunctions (see Fig. \ref{fig:data_avail}). Stations that exhibited prolonged gaps for both $T_a$ and $RH$ during the first year, on the other hand, were removed at comparably early points, as is visible in the cases of FRSEEP (removal rank = 12), FRASHA (13), and FRTECH (17).

\section{Discussion}
\label{Discussion}
\subsection{Imputation accuracy}
In this study, we developed a modeling framework and station removal procedure to systematically investigate the effects of urban WSN density reduction on predictive accuracy for $T_a$ and humidity, expressed as $RH$ and $e$. 
With mean prediction RMSEs between 0.69-0.93 K ($T_a$), 3.84-4.88\% ($RH$), and 0.72-0.9 hPa ($e$) for {EGB  1$\rightarrow$2} models containing between 42 (full WSN) and 2 stations, substantial thinning of the examined WSN could be achieved while maintaining high imputation accuracy. The largest RMSE increases occurred in the later stages of WSN density reduction. For instance, models using seven stations --- corresponding to a reduction of more than 80\% --- only showed an increase in prediction errors by 0.09 K (+13\% relative to the full model), 0.29\% (+10\%), and 0.07 hPa (+10\%) for $T_a$, $RH$, and $e$, respectively.
The effects of WSN reductions on RMSE are comparable to those reported by both \cite{honjoNetworkOptimizationEnhanced2015} and \cite{dingMachineLearningassistedMapping2023}, with RMSE increases of approximately 0.08 K, 0.3 K, and 0.48 K for station density reductions of 10\%, 70\%, and 90\% reported by \cite{honjoNetworkOptimizationEnhanced2015}, and 0.3 K, 0.6 K, 1.2 K for reductions of 30\%, 60\%, and 90\% by \cite{dingMachineLearningassistedMapping2023} (both studies were restricted to $T_a$).

While the overall predictive performance of the {EGB  1$\rightarrow$2} models was satisfactory, we found substantial differences between stations in different settings: stations located centrally and in the built-up parts of the city could generally be imputed accurately, whereas errors for stations in non-built-up locations were higher. This suggests that imputation at forest locations is generally challenging due to the distinct spatio-temporal dynamics of $T_a$ and humidity compared to urban areas. Consequently, we argue that such stations should not be removed if accurate data at their respective locations is essential for the study objective at hand.

Moreover, we found systematic biases in the predictions for the second study year for several stations, most notably in the aforementioned forest locations as well as for stations FRWSEE and FRVAUB, both of which are located in built-up areas in the east and the south of the city, respectively. Given that the second study year was, on average, warmer than the first year, the negative biases found at all affected stations except FRVAUB could partly be explained by interannual temperature differences. However, contrasting $T_a$ patterns between the two study years do not explain why, of all the stations in built-up areas, only FRVAUB and FRWSEE were affected, and why the biases for FRVAUB were positive. In light of this, and considering the variability between affected stations in the temporal development of the biases, it appears reasonable to assume that equipment-related issues such as sensor drift are at least partly responsible for the found prediction biases.

Despite differences in prediction accuracy between stations, even {EGB  1$\rightarrow$2} models trained on only a few predictor stations showed good overall performance across different meteorological conditions, indicated by the accurate representation of elevated $T_a$ measurements and modeled frequency climatic indicator days. While most of these were underestimated to some extent, the absolute deviations were mostly small and are to be expected given that the chosen climatic indicator days typically represent extreme conditions and are consequently only sparsely represented in the training data. Moreover, since climatic indicator days were defined based on exceeding or falling below fixed threshold values, even slight deviations of modeled values could lead to a changed estimated frequency of such days, particularly for low total numbers of observed events, as was the case for desert days, ice days, and tropical nights.

While prediction errors of $T_a$ rarely exceed 2 K in either direction around the observed value, large errors exceeding -12 K and +8 K were occasionally found in the {EGB  1$\rightarrow$2} models for stations located at the outer edges of the study area. Analysis of the corresponding periods revealed that these errors coincided with sudden temperature changes due to passing weather fronts, which were first registered at stations positioned furthest in the wind direction and initially missed by WSN subsets not containing these stations. This limitation implies that thinned WSNs are less suited for providing real-time estimates at former measurement sites and more appropriate for reproducing long-term spatial temperature patterns.

\subsection{Differences between model variants}
While the worst performance of the numerical model {SUEWS} compared to the EGB variants and especially the comparatively simple {GLMs} trained on the two DWD stations was surprising, there are several possible explanations for this finding: First, {SUEWS} produced estimates for 500 m $\times$ 500 m grid cells, whereas the {EGB} and {GLM} models were optimized to reproduce point measurements. While measurement locations were selected to be representative of their respective neighborhoods, these neighborhoods are not necessarily identical to the grid cells used by {SUEWS}, potentially leading to differences in the areas represented by the models' predicted values. Furthermore, considerable micro-scale variations in $T_a$ are possible even within areas featuring a single LCZ \citep{quanz2018micro}. 
Second, due to computational constraints, {SUEWS} was run in offline mode for this study, that is, it was not coupled with a mesoscale weather model, which can be expected to affect model errors, especially for topographically complex cities such as Freiburg.

Regarding the benefits of the guided station removal procedure compared to randomly chosen station subsets, we only observed substantial differences at high degrees of WSN density reduction. This is in contrast to the results of \cite{honjoNetworkOptimizationEnhanced2015}, who found small but largely constant differences between random and guided station sampling procedures (clustering based on geographic coordinates in their case) of 0.01-0.03 K across the entire range of sampling ratios. This discrepancy in the importance of guided station selection appears to be linked to initial station density. Our full WSN had an initial density of approximately 4.5 km\textsuperscript{2} per station, as opposed to 50 km\textsuperscript{2} per station for the Tokyo WSN used by \cite{honjoNetworkOptimizationEnhanced2015}. As progressive removal of stations brought our effective station density closer toward that of \cite{honjoNetworkOptimizationEnhanced2015}, the relative performance differences between random and guided sampling procedures converged as well. Therefore, we conclude that a guided selection becomes increasingly beneficial and important as the area covered by each station approaches and exceeds 50 km\textsuperscript{2}. In addition to spatial station density, the absolute number of remaining stations is likely to be significant too, as for higher station numbers, individual stations can be substituted by other stations exhibiting similar characteristics, whereas for smaller WSN subsets, each remaining station offers essential information and cannot be removed without substantial increases in modeling errors.

The largest differences in model performance between EGB  variants were found between the {EGB  1$\rightarrow$2} models, which performed temporal extrapolation (generating predictions for periods not included in training data), and the {EGB  1$\rightarrow$1} and {EGB  1,2$\rightarrow$1,2} models, where training and evaluation periods were identical. In practical terms, these results correspond to the accuracy expected when operating a recently thinned-out WSN to generate estimates for all initial measurement locations ({EGB  1$\rightarrow$2}) versus merely filling gaps in an unchanging WSN ({EGB  1$\rightarrow$1} and {EGB  1,2$\rightarrow$1,2}). Our results show that this kind of temporal extrapolation increases imputation errors, although the EGB  1$\rightarrow$2 models still produced largely accurate estimates even at high station removal rates, particularly considering that those stations exhibiting the highest errors could be retained to minimize the need for imputation at these 

When thinning a WSN in practice, one would hope to be able to estimate in advance the size of the modeling error after station removal (temporal extrapolation) and its development with increasing thinning rates based on the characteristics of the station's surroundings and its imputation errors while the station is deployed (gap-filling). Based on our findings, it appears that at remote forest locations, temporal extrapolation substantially increased prediction errors compared to gap-filling (Figs. \ref{fig:rmse_no_stations_ta} and \ref{fig:rmse_no_stations_rh}), making subsequent WSN thinning effects marginal. 
For remote open stations and more centrally located forest stations, gap-filling errors were often high, but temporal extrapolation and WSN thinning increased errors only slightly to moderately. 
For central stations, however, both the cost of temporal extrapolation and of reducing the WSN density exhibit substantial variability with no obvious explanations. Future research should therefore a) analyze when and under which meteorological conditions each central station exhibits high or low prediction errors to identify systematic patterns that might be related to station surroundings and b) apply our methodology to other WSNs to investigate whether the variability for central stations is also present in urban areas with different topographic characteristics and background climates.

Based on the generally small differences in performance between the {EGB  1$\rightarrow$1} and {EGB  1,2$\rightarrow$1,2} models, it appears that WSN operation times beyond one year provided little additional benefit in terms of gap-filling for non-functional stations. A possible explanation for this finding is that temporally variable effects of synoptic weather conditions (e.g., different rates of atmospheric mixing) and their interactions with urban morphology (e.g., cumulative heat storage by urban fabrics during a heat wave) cause variability among the spatial patterns of $T_a$ at different times \citep{beck2018air}, which cannot be distinguished by the models as they only have access to the current measurements of $T_a$ and $e$. Including information on the current and recent synoptic weather conditions could enable the models to learn more complex patterns and, therefore, make longer deployment times more beneficial.

\subsection{Station removal order and spatial patterns of WSN subsets}

The station removal procedure implemented in this study produced distinct spatial patterns. Stations at the outer edges of the WSN, particularly those situated in forests, were consistently removed first. This pattern indicates that both the geographic isolation of edge stations and the distinct micro-meteorology of forest environments limit their utility for imputing data at other locations.
It is important to note, however, that the early removal of these stations is closely linked to the way in which the effect of removing a station is calculated. Here, we chose to weigh all stations equally when quantifying the effect of removing any single station, by always imputing every station at every time step of the respective test data set. As remote forest stations a) always exhibited high imputation errors, regardless of whether they were part of the predictor stations, and, b) did not provide much useful information regarding the imputation of data from other stations, they were consistently removed first. In practice, only stations that are no longer part of the WSN would need to be permanently imputed, whereas stations that remain operational would only require imputation for periods in which they are non-functional. Consequently, removed stations could be assigned greater weight when assessing the effect of station removal on overall imputation accuracy, which would result in primarily retaining stations that are difficult to impute. In this study, the focus was to identify station subsets that minimize imputation error across all stations. For future applications, however, the presented algorithm can be flexibly adjusted to optimize station removal based on arbitrary criteria, including the aforementioned retention of high-error stations, but also minimizing errors for particular stations, or during periods of special interest, such as hot nights and heat or cold waves. 

In the intermediate removal stages, there is a larger amount of variability in the removal order of many central and suburban stations between CV folds. This can be seen as further evidence that, for higher WSN densities and station numbers, different station subsets can achieve similar imputation accuracy, and explains why models trained on random station subsets exhibited a virtually identical performance to those using selected station subsets for models containing ten or more predictor stations.

The fact that FRLAND and FRLORE were consistently retained longest suggests that strategically positioned stations at opposing edges of the urban area are optimal for reproducing city-wide spatial patterns of $T_a$ and humidity. This effectiveness probably stems from two key factors: First, their edge locations provide mixed source area signals encompassing both built-up surfaces and open, rural environments, facilitating accurate predictions across diverse urban settings. Second, positioning retained stations at opposite ends of the city minimizes the distance of any removed station to its nearest retained station, thereby enabling the models to take advantage of the spatial autocorrelation present in $T_a$ and humidity distributions. In that context, it has to be noted, however, that in the case of two predictor stations, {GLM}s performed marginally better than the {EGB  1$\rightarrow$2} models, which can be seen as evidence that for such a low number of predictor stations, the simpler parametric models are more suitable than the comparably complex, non-parametric EGB models.

\subsection{Limitations}

The methodology presented in this study has several important limitations: First, while the EGB  algorithm scales well to large input data sets \citep{xgboost}, the step-wise station removal algorithm becomes computationally intensive when applied to larger WSNs containing hundreds of stations. In such cases, we recommend increasing the number of stations removed at each step or combining our approach with additional pre-selection techniques, such as clustering based on geographic position or micro-climatic settings. 

Second, our modeling approach only produces estimates for specific locations for which previous stationary measurements are available and can currently not be used to obtain gridded city-wide predictions. However, our method can be adapted to produce spatial grids of predictions by adding relevant spatial predictors (such as land cover factions or urban morphometric parameters) during the training phase and subsequently using the learned relationships to generate estimates for non-sampled locations.

Lastly, our approach is not suitable for modeling other selected parameters, such as radiation fluxes and wind speed at pedestrian level, both of which are required for assessing thermal comfort, due to the higher spatio-temporal variability of these variables compared to $T_a$ and $RH$ in complex urban environments. To obtain these variables at high resolution and for longer periods, successful alternative approaches exist (e.g., \cite{briegel2023modelling}), which can be combined with our modeling framework for $T_a$ and $RH$. Nonetheless, even data on only $T_a$ and $RH$ provide valuable information as they are particularly relevant for thermal stress during the night \citep{oleson2015interactions} and in hot and humid climates \citep{coffel2017temperature, raymond2020emergence}.

\subsection{Conclusion}

In this study, we proposed, developed, and tested a step-wise station removal algorithm. We explored the effect of the algorithm applied to an urban WSN in Freiburg, Germany, on the imputation accuracy of $T_a$ and humidity. Based on our findings, we draw the following conclusions:

Urban WSNs can be substantially thinned after a one-year deployment period, utilizing learned spatial patterns to reproduce data at all initial measurement locations during subsequent periods.

Predictive performance varied systematically by station location and surrounding land cover. Remote forest stations consistently exhibited higher errors compared to stations in built-up but also in rural but open locations, likely due to more predictable spatial relationships and reduced microclimatic complexity at the open rural sites.

WSN reduction strategies should combine value-based selection with error-based retention. Based on our results, we recommend retaining 10-25\% of the most valuable stations for reproducing the full network's spatial patterns, supplemented by individual stations that cannot be modeled accurately, which can be identified through high imputation errors during the initial deployment period.

For WSNs in small or medium-sized cities and exhibiting a high spatial density ($<$10 km\textsuperscript{2} per station), differences between random and guided station selection procedures regarding overall imputation accuracy are minor. However, the choice of which stations to keep becomes important for lower spatial densities ($>$40 km\textsuperscript{2} per station) and lower absolute numbers of retained stations.

Operating reduced networks enables substantial reductions in operational costs and staff requirements while concentrating resources on fewer, strategically selected stations. Thinned WSNs offer a cost-effective alternative to numerical modeling for urban climate monitoring, allowing for the retention of stations for evaluating models and model output statistics at strategically relevant locations within a city. They further provide near-real-time data on strategic locations. In the future, this approach could also lead to an adaptive measurement approach where WSNs evolve in an iterative network configuration over time, keeping the relevant information at key stations while redeploying decommissioned stations in new geographic areas. \newline

\noindent \textbf{Acknowledgements}
\label{Acknowledgements}\newline
We gratefully acknowledge the valuable contributions of Ferdinand Briegel, who provided the SUEWS modeling for the numerical reference model, and of Simon Schrodi and Sina Farhadi, who offered valuable methodological advice. 
\newline

\noindent \textbf{Funding}\label{Funding}
The design, implementation, operation, and analysis of the station network received funding from the European Research Council Grant (ERC grant ``urbisphere - coupling dynamic cities and climate'' - agreement 855005). The development of the machine-learning algorithms received funding from the German Federal Ministry for the Environment, Nature Conservation, Nuclear Safety and Consumer Protection as part of the project ``I4C Intelligence for Cities'' (grant 67KI2029A/B) and from the German Science Foundation (DFG) as part of CRC 1537 ``Ecosense''.

\bibliographystyle{plainnat}
\bibliography{references}

\clearpage

 \appendix

\section{Additional figures and tables}

\begin{figure}[ht!]
 \centerline{\includegraphics[width=33pc]{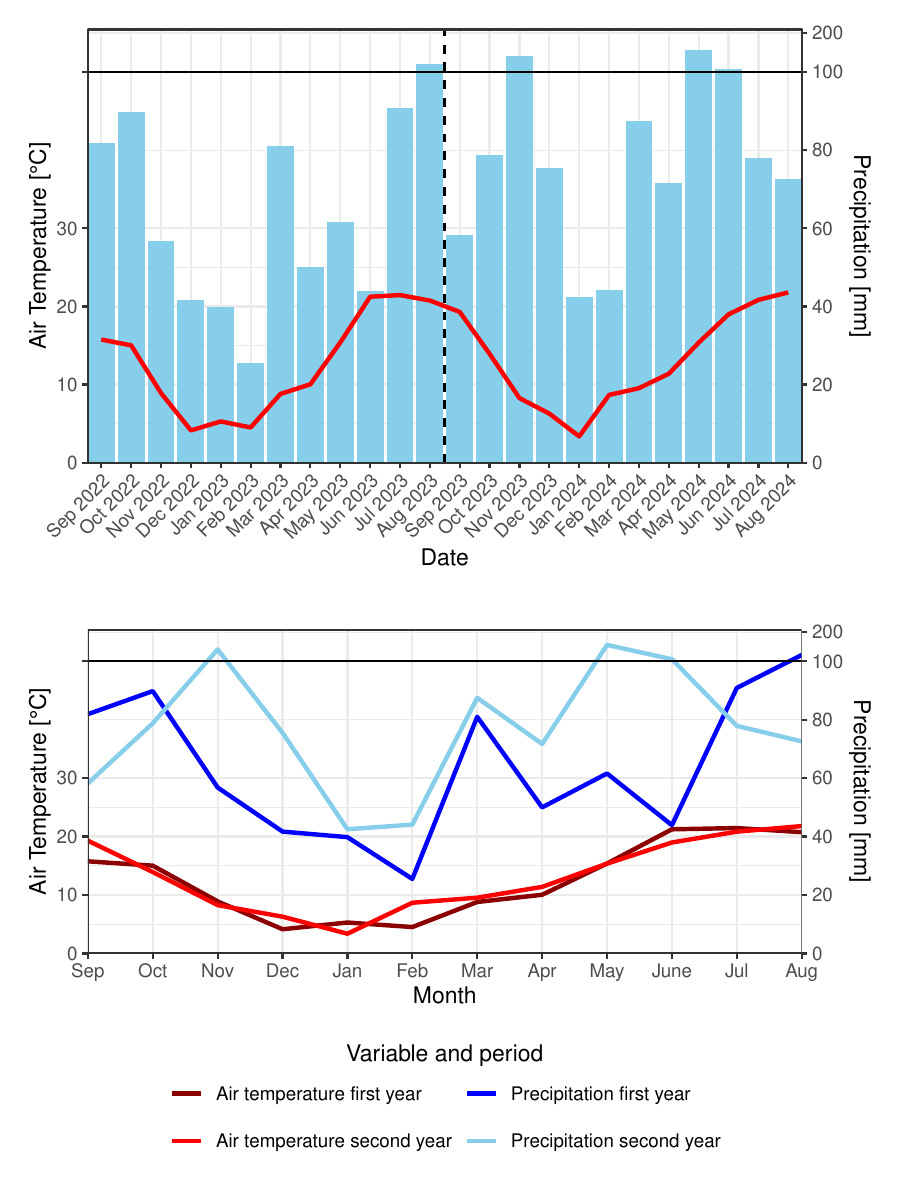}}
  \caption{Climate during the study period, displayed continuously (top) and month by month to enable comparison between the first and second year (bottom). Note the irregular scaling of the secondary axes in both figures.}
    \label{fig:combined}
\end{figure}

\begin{table}[]
\centering

\newsavebox{\mytablebox}
\savebox{\mytablebox}{%
\resizebox*{!}{0.85\textheight}{%
\begin{tabular*}{\textwidth}{@{\extracolsep{\fill}}lllllll@{}}
\toprule
ID     & Type & Elevation (m a.s.l.) & UAC   & LCZ & Longitude (°E) & Latitude (°N) \\ \midrule
FRASHA & T1   & 296                        & 11210 & 6   & 7.870209      & 47.986557    \\
FRBETZ & T2   & 250                        & 11210 & 5   & 7.817673      & 48.004900    \\
FRBRUH & T1   & 238                        & 12100 & 8   & 7.854189      & 48.031009    \\
FRDIET & T2   & 230                        & 13300 & D   & 7.792878      & 48.013766    \\
FRDREI & T1   & 260                        & 12220 & 6   & 7.826545      & 47.995872    \\
FREBNE & T1  & 340                        & 23000 & D   & 7.922671      & 47.983431    \\
FREICH & T2   & 695                        & 31000 & A   & 7.882986      & 47.954049    \\
FRFRIE & T2   & 257                        & 14100 & 9   & 7.841538      & 48.010887    \\
FRGART & T2   & 262                        & 11220 & 6   & 7.824259      & 47.986798    \\
FRGLBA & T2   & 289                        & 11220 & 6   & 7.868736      & 48.006918    \\
FRGUNT & T1   & 339                        & 11210 & 6   & 7.858853      & 47.964012    \\
FRHAID & T2   & 235                        & 12100 & 8   & 7.785271      & 47.985761    \\
FRHBHF & T2   & 271                        & 12230 & 5   & 7.842381      & 47.999806    \\
FRHERD & T1   & 266                        & 11210 & 6   & 7.856615      & 48.006103    \\
FRHOCH & T2   & 227                        & 21000 & D   & 7.805014      & 48.053762    \\
FRHOLZ & T2   & 279                        & 11210 & 5   & 7.850207      & 47.990968    \\
FRIHOC & T2   & 223                        & 12100 & 8   & 7.816336      & 48.036945    \\
FRINST & T2   & 274                        & 11210 & 5   & 7.848589      & 48.000045    \\
FRKART & T2   & 279                        & 11100 & 2   & 7.850794      & 47.996988    \\
FRLAND & T1   & 234                        & 11210 & 5   & 7.804825      & 48.019458    \\
FRLORE & T2   & 281                        & 11210 & 6   & 7.845321      & 47.981176    \\
FRMERZ & T1  & 311                        & 23000 & B   & 7.827334      & 47.960125    \\
FRMESS & T2   & 249                        & 12100 & 8   & 7.842709      & 48.016982    \\
FROPFS & T2   & 217                        & 50000 & G   & 7.764530      & 48.012593    \\
FROWIE & T2   & 284                        & 11210 & 6   & 7.856149      & 47.986906    \\
FRPDAS & T1   & 279                        & 12100 & 5   & 7.845668      & 47.995162    \\
FRRIES & T1   & 237                        & 11100 & 6   & 7.791470      & 47.998112    \\
FRSEEP & T2   & 248                        & 14100 & B   & 7.819404      & 48.011264    \\
FRSTGA & T2   & 279                        & 14100 & B   & 7.857968      & 47.997815    \\
FRSTGE & T2   & 244                        & 11210 & 6   & 7.799452      & 47.980250     \\
FRSTUH & T2   & 269                        & 11100 & 5   & 7.835886      & 47.995906    \\
FRTECH & T1  & 247                        & 11230 & 6   & 7.829959      & 48.015290     \\
FRTIEN & T1   & 211                        & 21000 & D   & 7.723827      & 47.981533    \\
FRUNIK & T2   & 263                        & 11210 & 5   & 7.839931      & 48.005184    \\
FRUWIE & T2   & 272                        & 11220 & 5   & 7.842559      & 47.988523    \\
FRVAUB & T1   & 259                        & 11210 & 6   & 7.823044      & 47.976285    \\
FRWAHS & T2   & 324                        & 31000 & A   & 7.840420      & 47.970834    \\
FRWEIN & T2   & 248                        & 11210 & 4   & 7.807998      & 47.995228    \\
FRWILD & T2   & 293                        & 23000 & 9   & 7.892956      & 48.032503    \\
FRWITT & T2   & 444                        & 23000 & 9   & 7.821302      & 47.937864    \\
FRWSEE & T2   & 315                        & 11210 & 6   & 7.893637      & 47.985970     \\
FRZAHR & T2   & 277                        & 11210 & 6   & 7.866976      & 48.020498    \\ \cmidrule(r){1-7}
\end{tabular*}
}
}

\begin{minipage}{\wd\mytablebox}
\caption{Overview of station characteristics. T1\textsuperscript{1} refers to T1 stations that do not feature a Black Globe sensor. UAC and LCZ refer to the Urban Atlas Classes (\citealp{Montero2014}, data derived from \citealp{european_env_agency_2018}) and Local Climate Zones (\citealp{stewart2012local}, assigned manually based on \citealp{matthias_demuzere_2023_8419340}). Elevation data were obtained from \citet{lgl}.}
\label{tab:stations_overview}
\centering
\usebox{\mytablebox}
\end{minipage}
\end{table}

\begin{figure}[ht!]
 \centerline{\includegraphics[width=39pc]{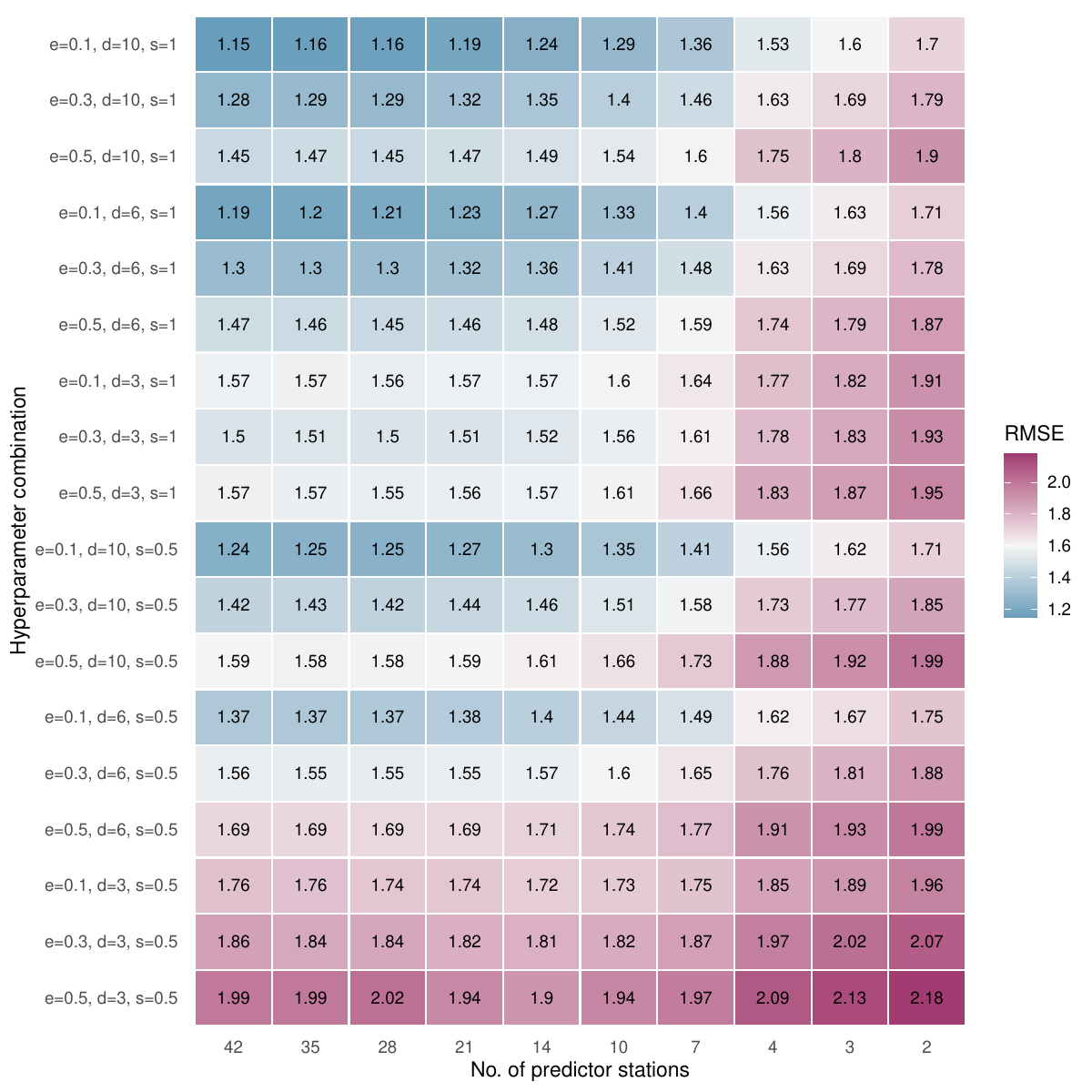}}
  \caption{Results of hyperparameter optimization for different numbers of predictor stations. e = eta (learning rate), d = max depth (maximum tree depth), s = subsample (subsampling of training instances). RMSEs are averaged across stations and variables (scaled between 0 and 1) and multiplied by 100 to reduce the number of required digits displayed.}\label{fig:hyp_opt}
\end{figure}

\begin{figure}[ht!]
\centerline{\includegraphics[width=39pc]{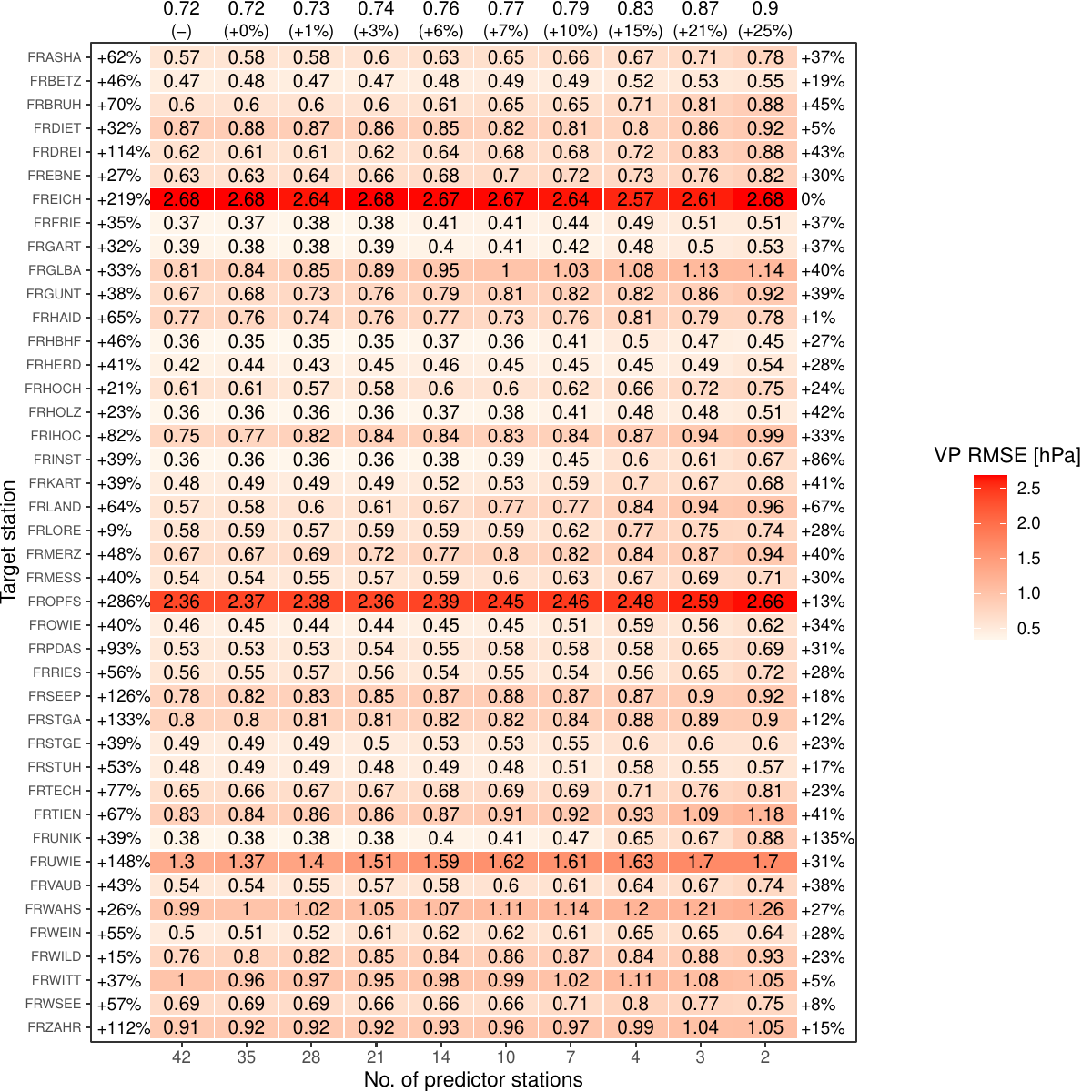}}
  \caption{Station-wise prediction RMSEs of \textit{EGB  1-1} models across the number of used predictor stations for $e$. Otherwise identical to Fig. \ref{fig:rmse_no_stations_ta}}
  \label{fig:rmse_no_stations_vp}
\end{figure}

\begin{figure}[ht!]
\centerline{\includegraphics[width=39pc]{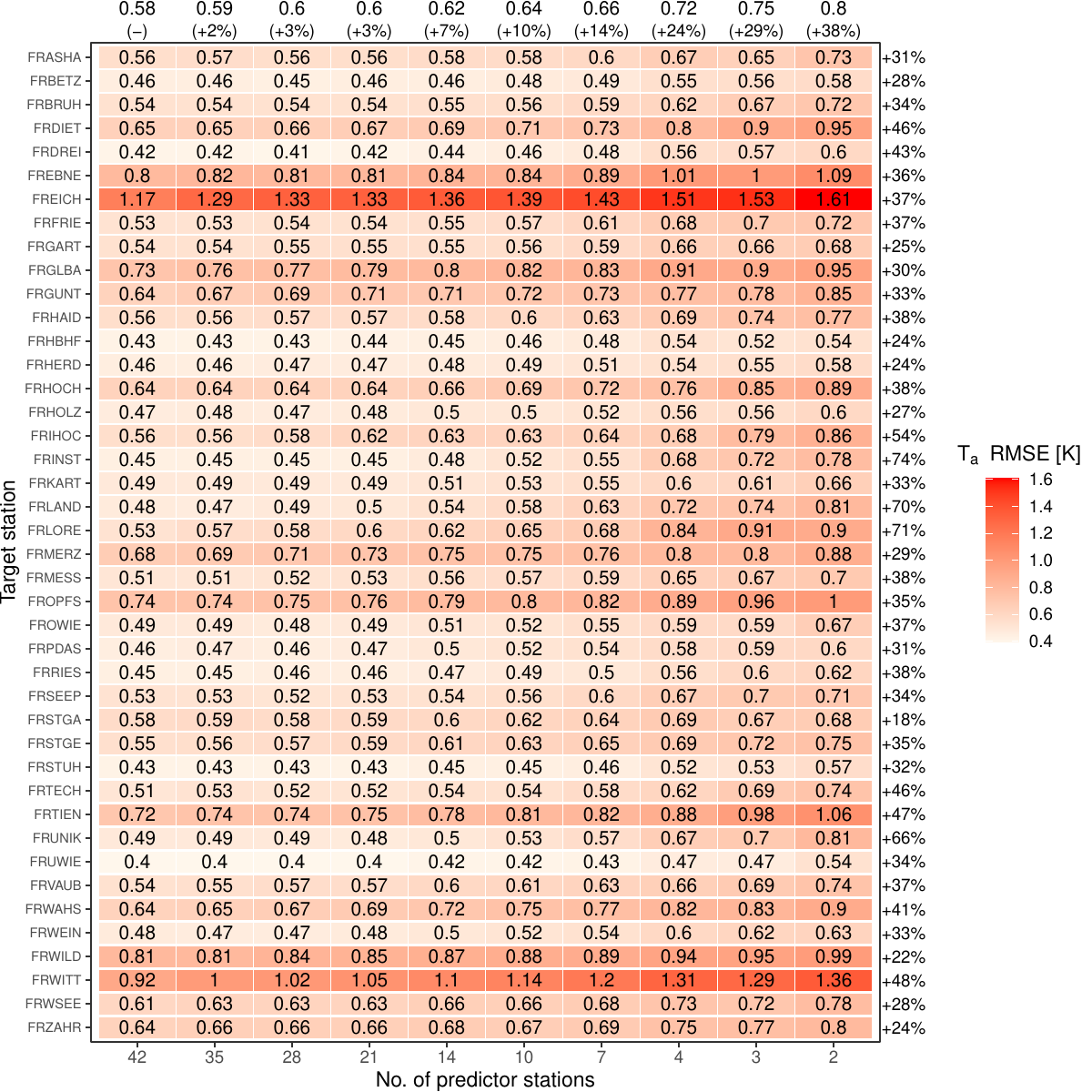}}
  \caption{Station-wise prediction RMSEs of \textit{EGB  1-1} models across the number of used predictor stations for $T_a$. Otherwise identical to Fig. \ref{fig:rmse_no_stations_ta}.}
  \label{fig:rmse_no_stations_ta_first_year}
\end{figure}

\begin{figure}[ht!]
\centerline{\includegraphics[width=39pc]{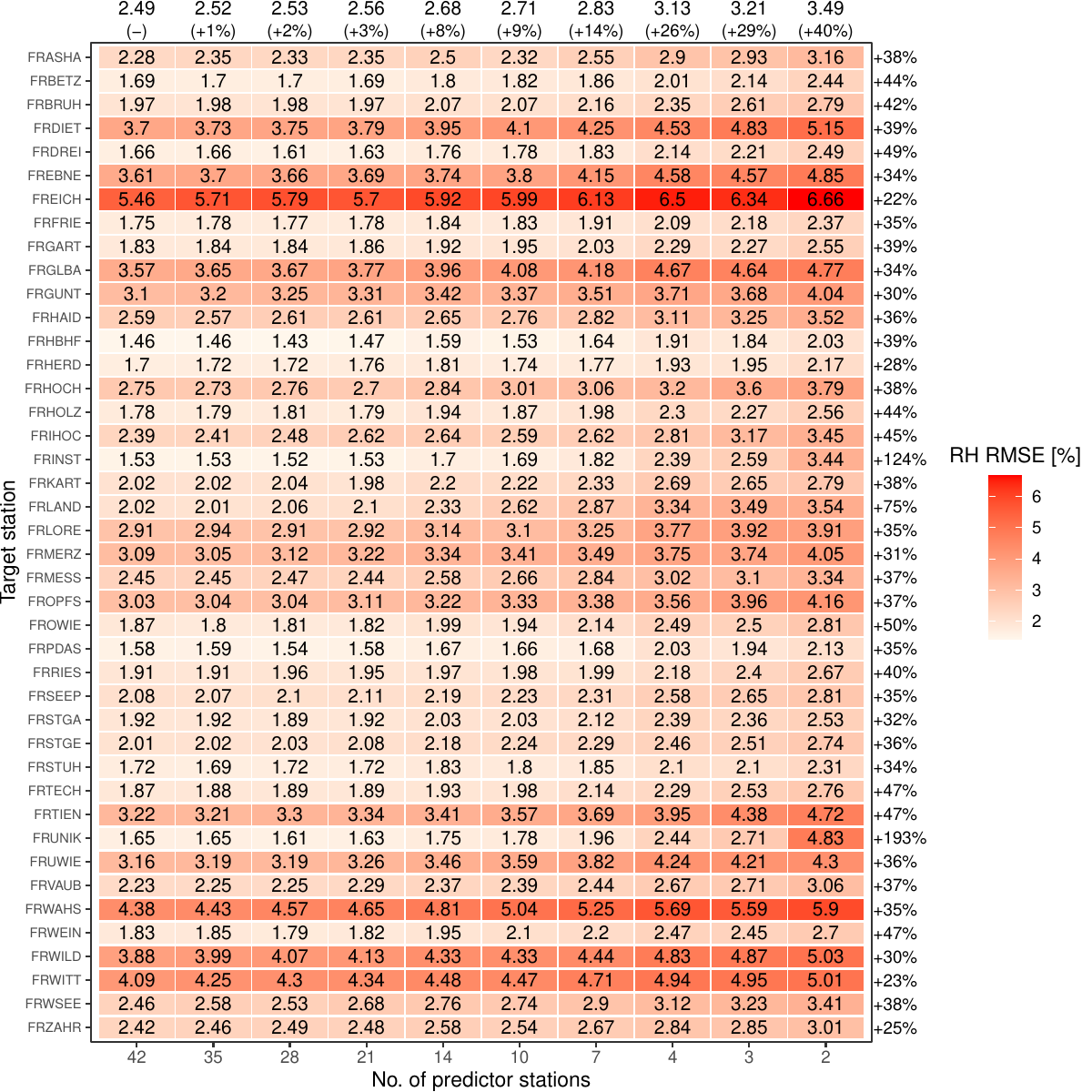}}
  \caption{Station-wise prediction RMSEs of \textit{EGB  1-1} models across the number of used predictor stations for $RH$. Otherwise identical to Fig. \ref{fig:rmse_no_stations_ta}.}
  \label{fig:rmse_no_stations_rh_first_year}
\end{figure}

\begin{figure}[ht!]
\centerline{\includegraphics[width=39pc]{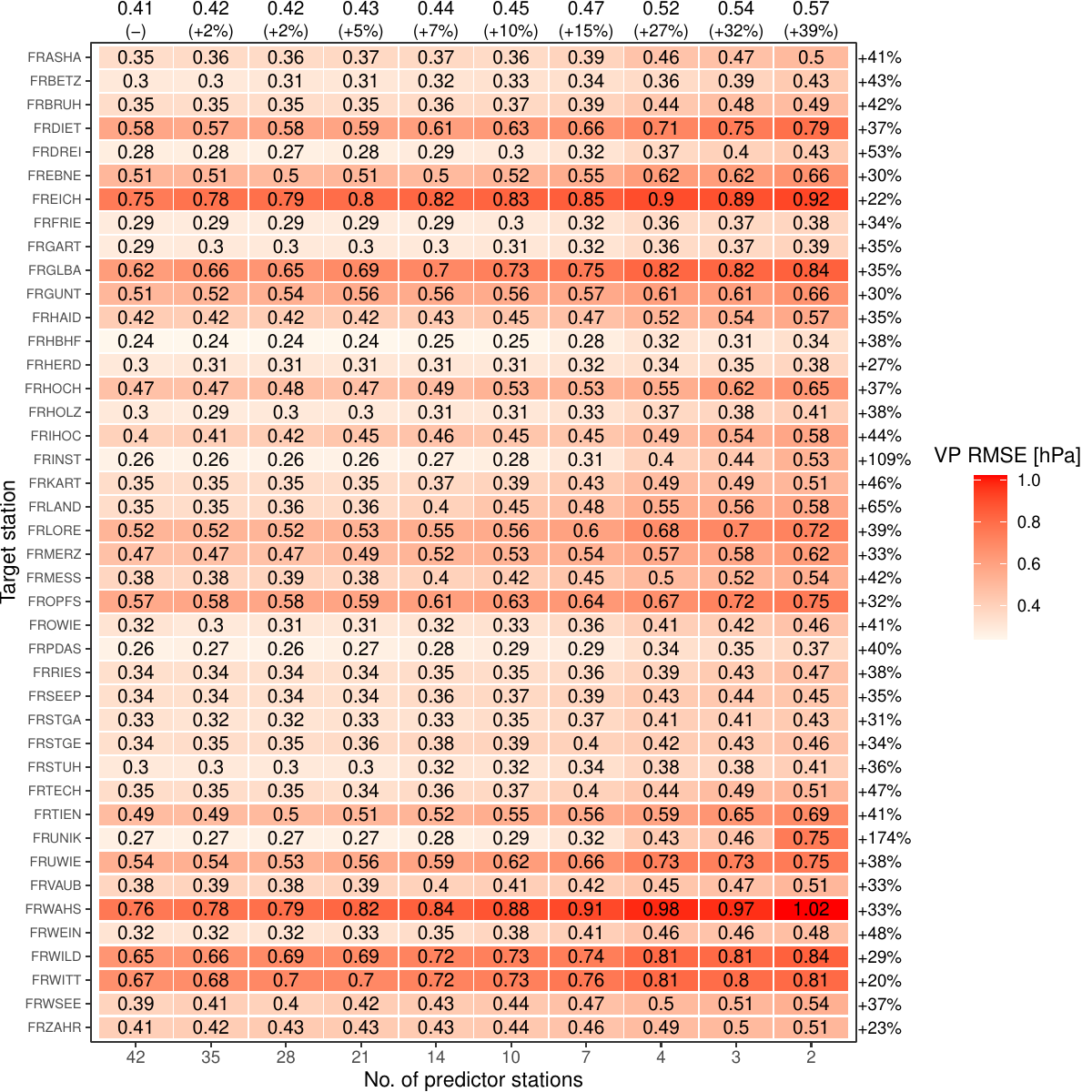}}
  \caption{Station-wise prediction RMSEs of \textit{EGB  1-1} models across the number of used predictor stations for $e$. Otherwise identical to Fig. \ref{fig:rmse_no_stations_ta}.}
  \label{fig:rmse_no_stations_vp_first_year}
\end{figure}

\begin{table}[ht!]
\centering
\caption{Criteria for climatic indicators.}
\label{tab:clim_stats_crit}
\begin{tabular}{lll}
\toprule
Statistic      & Criteria       & Time\\ \midrule
Summer day     & max($T_a$) $\geq$ 25ºC  & 0-23 UTC    \\
Hot day        & max($T_a$) $\geq$  30ºC  & 0-23 UTC    \\
Desert day     & max($T_a$) $\geq$  35ºC  & 0-23 UTC    \\
Frost day      & min($T_a$) $<\,\,$  0ºC   & 0-23 UTC    \\
Ice day        & max($T_a$) $<\,\,$  0ºC   & 0-23 UTC    \\
Tropical night & min($T_a$) $\geq$ 20ºC & 18-06 UTC   \\ \bottomrule
\end{tabular}
\end{table}

\begin{table}[]
\centering
\savebox{\mytablebox}{%
\resizebox*{!}{0.93\textheight}{%
\begin{tabular}{@{}ccccccc@{}}
\toprule
Station ID & \multicolumn{2}{c}{Ta [K]}   & \multicolumn{2}{c}{RH [\%]}   & \multicolumn{2}{c}{e [hPa]}   \\
           & First year & Second year & First year & Second year & First year & Second year \\ \midrule
FRASHA     & -0.01       & 0.02       & 0.03      & -1.35        & 0.01      & -0.30        \\
FRBETZ     & 0.00       & 0.06       & 0.07      & 0.46       & 0.00       & 0.05       \\
FRBRUH     & 0.01      & -0.02        & 0.11      & -1.16        & 0.00       & -0.30        \\
FRDIET     & 0.00       & 0.08       & 0.05      & 3.89       & 0.01       & 0.40       \\
FRDREI     & 0.02      & 0.05       & 0.00       & -1.88        & -0.01       & -0.39        \\
FREBNE     & 0.04      & 0.08       & 0.25      & -0.47        & 0.02       & -0.13        \\
FREICH     & 0.01      & -0.79        & 0.22      & -8.00        & 0.05       & -1.47        \\
FRFRIE     & -0.01       & 0.01       & 0.02      & 0.56       & 0.01       & 0.08       \\
FRGART     & -0.01       & 0.00        & -0.04       & -0.14        & 0.00       & -0.02        \\
FRGLBA     & 0.03      & 0.09       & -0.09       & -0.54        & -0.02       & -0.15        \\
FRGUNT     & 0.01      & 0.07       & 0.06      & -1.73        & 0.00       & -0.31        \\
FRHAID     & 0.00       & 0.01       & 0.03      & 2.54       & 0.01       & 0.44       \\
FRHBHF     & -0.02       & -0.06        & 0.03      & 0.78       & 0.01       & 0.15       \\
FRHERD     & 0.02      & 0.02       & 0.00       & -0.73        & 0.00       & -0.17        \\
FRHOCH     & 0.01      & 0.02       & 0.05      & 0.77       & 0.01       & 0.12       \\
FRHOLZ     & 0.00       & 0.01       & 0.04      & -0.21        & 0.01       & -0.01        \\
FRIHOC     & 0.00       & 0.04       & 0.09      & -0.93        & 0.02       & -0.21        \\
FRINST     & -0.02       & -0.04        & 0.06      & 1.29       & 0.01       & 0.21       \\
FRKART     & -0.03       & 0.00        & 0.07      & -0.06        & 0.03       & 0.04       \\
FRLAND     & 0.03      & 0.06       & 0.09      & -1.85        & 0.00       & -0.37        \\
FRLORE     & 0.02      & 0.00        & -0.01       & 1.62       & 0.00       & 0.26       \\
FRMERZ     & 0.03      & 0.11       & 0.10      & -0.49        & 0.00       & -0.19        \\
FRMESS     & -0.01       & -0.03        & -0.01       & 0.79       & 0.00       & 0.09       \\
FROPFS     & 0.01      & -0.72        & -0.06       & -4.48        & -0.01       & -1.16        \\
FROWIE     & 0.01      & 0.04       & 0.04      & 0.44       & 0.01       & 0.10       \\
FRPDAS     & 0.01      & 0.02       & 0.00       & -1.41        & 0.00       & -0.30        \\
FRRIES     & 0.01      & -0.03        & 0.07      & -1.09        & 0.01       & -0.25        \\
FRSEEP     & -0.01       & 0.15       & 0.09      & 0.18       & 0.02       & -0.06        \\
FRSTGA     & 0.00       & 0.02       & 0.05      & 0.13       & 0.01       & -0.05        \\
FRSTGE     & 0.00       & 0.02       & 0.03      & 0.84       & 0.01       & 0.15       \\
FRSTUH     & 0.01      & 0.01       & 0.05      & 2.01       & 0.01       & 0.31       \\
FRTECH     & 0.04      & 0.13       & 0.02      & -0.11        & -0.01       & -0.09        \\
FRTIEN     & 0.04      & 0.08       & 0.20      & -1.43        & 0.01       & -0.32        \\
FRUNIK     & -0.01       & -0.06        & 0.01      & -0.17        & 0.01       & 0.04       \\
FRUWIE     & 0.01      & 0.10       & -0.02       & -3.55        & -0.01       & -0.75        \\
FRVAUB     & 0.03      & 0.19       & 0.00       & 0.40       & -0.01       & -0.04        \\
FRWAHS     & 0.04      & -0.41        & -0.18       & -1.53        & -0.03       & -0.25        \\
FRWEIN     & -0.02       & -0.06        & 0.05      & 1.65       & 0.01       & 0.30       \\
FRWILD     & 0.00       & 0.00        & 0.07      & -0.61        & 0.00       & -0.12        \\
FRWITT     & -0.08       & 0.00        & 0.04      & 3.60       & 0.02       & 0.51       \\
FRWSEE     & -0.02       & -0.38        & 0.13      & 2.03       & 0.02       & 0.39       \\
FRZAHR     & 0.02      & 0.04       & -0.08       & -2.42        & -0.01       & -0.45        \\ \bottomrule
\end{tabular}
}
}

\begin{minipage}{\wd\mytablebox}
\caption{Mean prediction biases for all stations and variables for the first and second study year.}
\label{tab:bias_all_vars}
\centering
\usebox{\mytablebox}
\end{minipage}
\end{table}

\begin{figure}[ht]
 \centerline{\includegraphics[width=33pc]{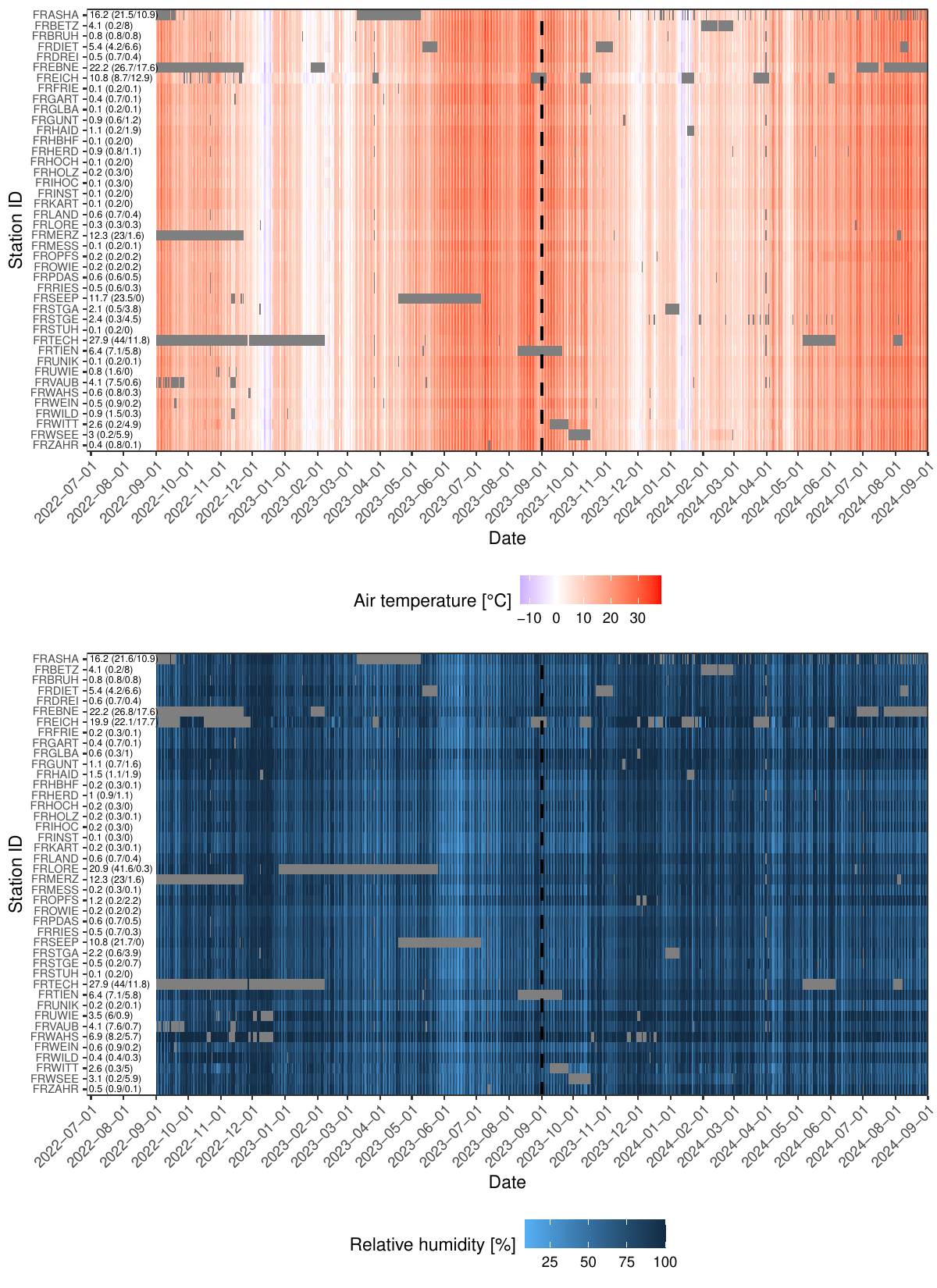}}
  \caption{Data availability across the study period for $T_a$ (top) and $RH$ (bottom). Grey areas indicate missing or deleted data. The values next to the station IDs give the percentage of missing data over the whole study period and separately for both study years. The black dashed line indicates the separation between the first and second study years.}\label{fig:data_avail}
\end{figure}

\end{document}